\pdfoutput=1

\documentclass[11pt]{article}

\usepackage[final]{acl}

\usepackage{times}
\usepackage{latexsym}
\usepackage{multirow}
\usepackage{booktabs}
\usepackage{amsmath}
\usepackage{amssymb}
\usepackage{adjustbox}
\usepackage{amsfonts}
\usepackage{arydshln}

\usepackage{cleveref}
\usepackage{graphicx}
\usepackage{epstopdf}
\usepackage{hyperref}
\usepackage{colortbl}
\usepackage{float}
\usepackage{pifont}
\usepackage{xcolor}
\usepackage{fontawesome5}
\definecolor{my_green}{RGB}{51,102,0}
\definecolor{my_red}{RGB}{204, 0, 0}
\renewcommand{\checkmark}{\textcolor{my_green}{\ding{51}}} %
\newcommand{\crossmark}{\textcolor{my_red}{\ding{55}}} %

\usepackage[T1]{fontenc}

\usepackage[utf8]{inputenc}

\usepackage{microtype}
\usepackage{makecell}

\usepackage{inconsolata}

\usepackage{graphicx}

\usepackage{dblfloatfix}

\title{Any Information Is Just Worth One Single Screenshot: Unifying Search With Visualized Information Retrieval}

\author{Ze Liu$^{1,2*}$, \ \ \ Zhengyang Liang$^{1,3*}$, \ \ \ Junjie Zhou$^{1,3*}$, \ \ \, Zheng Liu$^{1,4*}$\thanks{Core contributors, with Zheng Liu as the project lead.}, \ \ \  Defu Lian$^{2}$ \\
$^1$ BAAI, \ \ \ \ $^2$ USTC,  \ \ \ \ $^3$ BUPT,  \ \ \ \ $^4$ HKPU \\
\texttt{zhengliu1026@gmail.com} 
}

\begin{document}
\maketitle 

\begin{abstract}
With the popularity of multimodal techniques, it receives growing interests to acquire useful information in visual forms. In this work, we formally define an emerging IR paradigm called \textit{Visualized Information Retrieval}, or \textbf{Vis-IR}, where multimodal information, such as texts, images, tables and charts, is jointly represented by a unified visual format called \textbf{Screenshots}, for various retrieval applications. We further make three key contributions for Vis-IR. First, we create \textbf{VIRA} (Vis-IR Aggregation), a large-scale dataset comprising a vast collection of screenshots from diverse sources, carefully curated into captioned and question-answer formats. Second, we develop \textbf{UniSE} (Universal Screenshot Embeddings), a family of retrieval models that enable screenshots to query or be queried across arbitrary data modalities. Finally, we construct \textbf{MVRB} (Massive Visualized IR Benchmark), a comprehensive benchmark covering a variety of task forms and application scenarios. Through extensive evaluations on MVRB, we highlight the deficiency from existing multimodal retrievers and the substantial improvements made by UniSE. Our work will be shared with the community, laying a solid foundation for this emerging field. 
\end{abstract}

\begin{figure}[t]
    \centering
    \includegraphics[width=\linewidth]{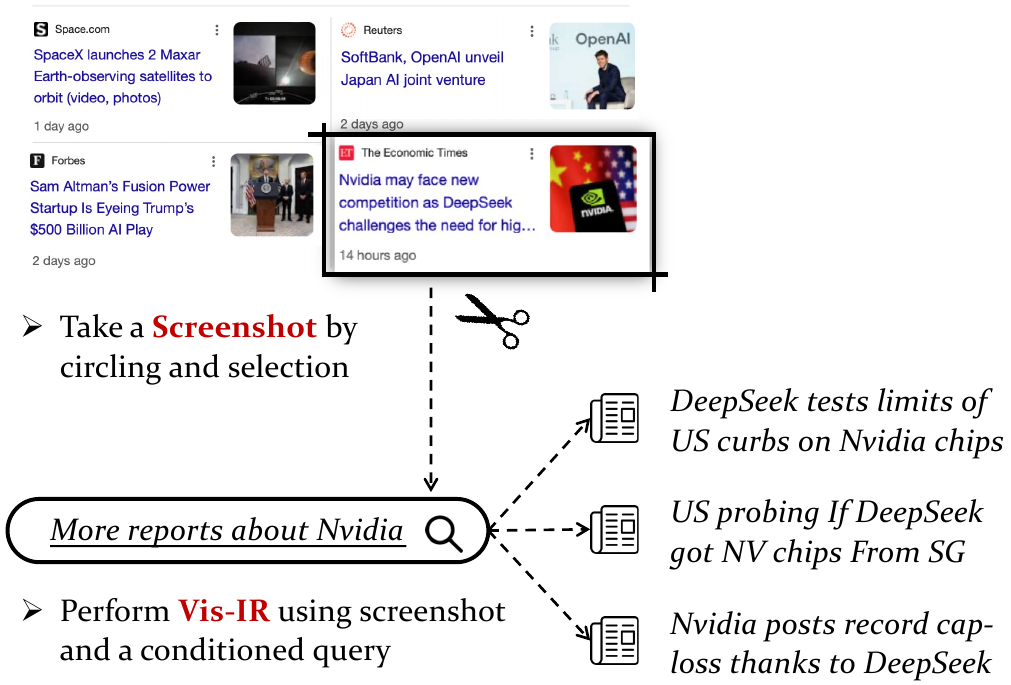}
    \vspace{-0.5cm}
    \caption{A use case of Vis-IR. Users take a screenshot of their interested news by circling and selection, and search for relevant news reports associated with ``Nvidia'' based on a query conditioned on the screenshot.} 
    \vspace{-0.5cm}
    \label{fig:example}
\end{figure}

\section{Introduction} 
Information retrieval has experienced tremendous progresses in recent years, driven by breakthroughs in foundation models. The growing capacity of language models has enabled precise and generalized text retrieval \cite{bge,wang2022text,e5-mistral}, while the development in vision-language models has extended information retrieval to a broader range of data modalities \cite{mbeir-wei2023uniir,magiclens,vista}. With the recent popularity of multimodal techniques \cite{blip2-li2023blip,llava-liu2023visual,team2023gemini}, there is growing interest in \textit{uniformly representing multimodal data in visual forms and leveraging these visual representations for diverse information retrieval tasks} (shown in Figure \ref{fig:example}) \cite{faysse2024colpali,ma2024unifying,zhang2024gme}. For example, researchers can circle and select a part of a paper, which may include texts, figures, equations, and figures, and search for relevant literature to address a specific question about the selected content. Similarly, people take a screenshot of an advertisement on their cellphones and retrieve descriptions of the corresponding product\footnote{\scriptsize https://blog.google/products/search/google-circle-to-search-android/}.
These innovative paradigms will greatly enhance the flexibility of information retrieval in real-world scenarios, leading to a unified paradigm for search engines. In this paper, we formally define these problems as \textit{Visualized Information Retrieval}, or \textbf{Vis-IR} for brevity. We also define the visual representation for a mixture of multimodal data as a \textbf{screenshot}, which is treated as a unified entity in the retrieval process. 

Despite that common multimodal retrievers can be applied for Vis-IR in a zero-shot manner, and there are preliminary works focused on improving representations of specific multimodal data, like screenshots of Wiki webpages \cite{ma2024unifying}, several fundamental challenges still persist for this emerging problem. Particularly, there are no tailored models which can offer unified, high-quality support for various Vis-IR applications. Besides, there is a lack of comprehensive benchmarks to evaluate a retriever's performance on Vis-IR tasks. Finally, no specialized datasets are curated and published for training competent Vis-IR models. 

To address the above challenges, we present the following three resources, which are critical to the technical advancement of Vis-IR. 

First, we create a large-scale dataset, namely \textbf{VIRA} (Vis-IR, Aggregation), which comprises a vast collection of 13 million screenshot data from diverse sources, like {webpages} from news websites, shopping platforms, and Wikipedia, homepages from GitHub, research papers from ArXiv, general PDF documents, and various forms of charts. Each screenshot is assigned with a \textit{fine-grained caption} by either extracting from meta data or annotating with sophisticated OCR tools. We further produce two types of \textit{question-answer data} for the well-captioned screenshots: 1) q2s tuples, which comprises a screenshot and a query about the screenshot, 2) sq2s triples, which contain source screenshots, conditioned queries about the source, and target screenshots. Altogether, over 20 million data samples are included by VIRA. 

Second, we develop a family of embedding models, called \textbf{UniSE} (Universal Screenshot Embeddings), built on top of our created dataset. These models enable screenshots to either serve as queries or be queried by various data formats. For example, using a screenshot to search for an image or a document, or retrieving a screenshot based on a textual question. We adopt two alternative model architectures for UniSE: one based on CLIP \cite{clip-radford2021learning}, which is more efficient, the other one leveraging MLLMs \cite{llava-liu2023visual}, which is more expressive. This allows users to flexibly choose the model that best suits their own needs. 

Third, we construct \textbf{MVRB} (Massive Visualized IR Benchmark) to evaluate multimodal retrievers' performance on general Vis-IR tasks. The benchmark includes various task types, such as screenshot-based multimodal retrieval (screenshot to anything, anything to screenshot) and screenshot-conditioned retrieval (e.g., searching for documents using queries conditioned on screenshots). It also covers a variety of important domains, including news, products, papers, and charts. We conduct a comprehensive experimental study using MVRB, which highlights significant deficiencies in existing methods and uncovers key factors influencing the development of Vis-IR retrievers. The results also demonstrate the effectiveness of UniSE in both in-domain and out-of-domain applications. 
 
To summarize, we define the Vis-IR problem, which holds promising prospects. We present three key resources: a large-scale dataset, powerful models, and a comprehensive benchmark, which are foundational to Vis-IR and will be shared with the public to facilitate technical advancement. 



\begin{figure*}[t]
    \centering
    \includegraphics[width=0.9\textwidth]{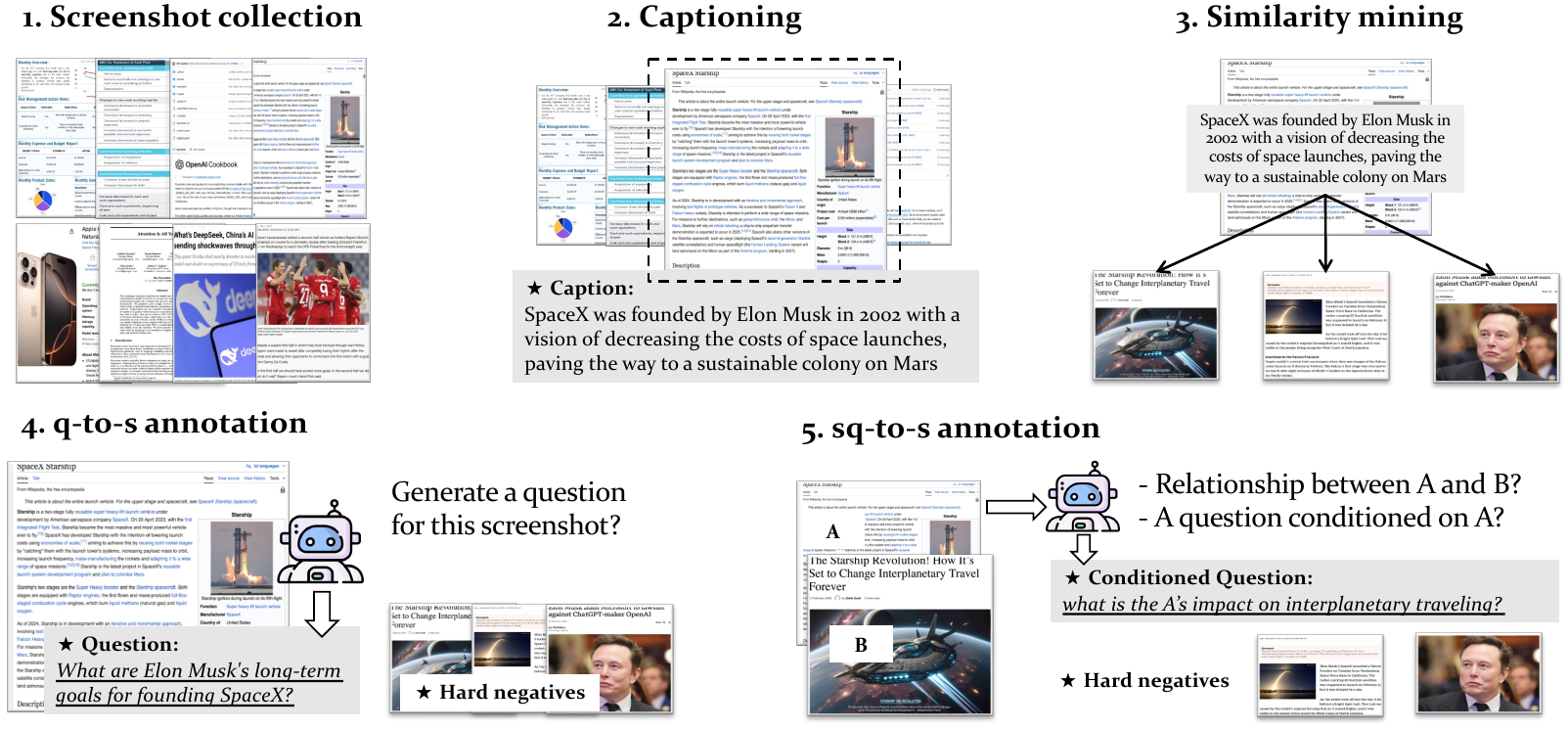}
    \vspace{-0.3cm}
    \caption{Creation process of VIRA dataset, including 1) comprehensive screenshot collection from various sources, 2) fine-grained screenshot captioning, 3) similar screenshots mining, 4) q2s annotation, and 5) sq2s annotation.} 
    \vspace{-0.3cm}
    \label{fig:data}
\end{figure*}

\section{Dataset: VIRA} 
In this section, we present the VIRA (Vis-IR Aggregation) dataset, which offers diverse and large-scale training data for developing Vis-IR models. While creating the dataset, we emphasize two main factors: 1. \textbf{comprehensiveness} of screenshots, 2. \textbf{utility} of annotation. The creation's outline is introduced as follows. Due to the space limitation, the entire details are provided in Appendix~\ref{appendix:details-VIRA}.


\subsection{Screenshot collection}
We perform massive collection of screenshot data spanning \textbf{seven categories}, which include \textit{News}: from popular websites like BBC, CNN, and Fox, \textit{Products}: from Amazon, \textit{Research Papers}: from ArXiv, \textit{General Documents}: from PDF Association, \textit{Charts}: from ArXiv, \textit{Common Knowledge}: from Wikipedia; \textit{Project Homepages}: from GitHub. The data is evenly distributed across these categories, leading to \textbf{13 million screenshots} in total. The collected data encompasses a rich variety of formats, showing diverse combinations of text, figures, and structures presented in various layouts. These features substantially enhance the generalization for models trained on corresponding data. 

\subsection{Screenshot Annotation}
We perform two types of annotations for the collected data. First, we assign each screenshot with a fine-grained \textbf{caption}, which paraphrases its detailed semantic. The caption is produced by in alternative ways. For those associated with complete meta data, e.g., project homepage from Github, we apply proper data extraction pipelines to acquire the captions. While for other free-form screenshots, we obtain captions based on OCR methods. Second, we create \textbf{question-answering} annotations for the well-captioned screenshots, which closely align with typical retrieval tasks. Specifically, we consider the following forms of question-answering data. 1. \textbf{q2s tuples}, denoted as (q, s). For a screenshot $s$, we prompt a large language model (LLM) to generate a question $q$ based on the screenshot's caption. For example, given a screenshot with the title “\textit{Nvidia posts record cap-loss due to DeepSeek}”, a question like “\textit{What’s the impact of DeepSeek on Nvidia?}” would be generated. 2. \textbf{sq2s triplets}, denoted as (s1, q, s2). Following MegaPairs methodology~\cite{zhou2024megapairs}, for a pair of relevant screenshots (s1, s2) mined from the corpus, we prompt the LLM to analyze their relationship based on their captions and generate a relational question. For example, if s1 is a breaking news screenshot and s2 is a subsequent news report, a question like “\textit{What’s the follow-up news to it?}” is generated. 
Finally, we create \textbf{13 millions} screenshot-caption samples and \textbf{7 millions} question-answering samples from the above operations. 

We supplement the above data with \textbf{hard negatives}, which are crucial for training retrieval models. For each $q2s$ tuple, we introduce hard negatives based on either textual or visual similarity (using off-the-shelf embedding models). For a $q2s$ tuple, a screenshot $s'$ is selected if it meets any of these conditions: \textit{1}) $s'$ has a similar caption with \textit{s}, \textit{2}) $s'$ is visually similar with \textit{s}, \textit{3}) $s'$ is textually similar with \textit{q}. While for a $s_1q$2$s_2$ triplet, we use $s'$ as a hard negative, if it enjoys a strong textual or visual similarity with the retrieval target $s_2$.

\section{Model: UniSE} 
We develop retrieval models based on VIRA dataset called Universal Screenshot Embeddings (UniSE), providing unified supports for general Vis-IR tasks. 

\subsection{Embedding}
We adopt two structures for UniSE, allowing users to make flexible selection based on their own needs. 

\paragraph{UniSE-CLIP.}
The first type adopts the \textit{CLIP architecture} \cite{clip-radford2021learning}, which is more time-efficient. In this form, the screenshot $s$ and text input $t$ are encoded by the visual and textual encoder of CLIP, respectively: $\boldsymbol{e}_s \leftarrow \phi_v(s)$, $\boldsymbol{e}_t \leftarrow \phi_t(t)$. For a composed input of a screenshot $s$ and conditioned textual query $q$, the joint representation is computed by linear combining the screenshot and query's embedding: $\boldsymbol{e}_{s,q} \leftarrow \boldsymbol{e}_s + \boldsymbol{e}_q$. Specifically, we utilize OpenAI's CLIP-Large\footnote{https://huggingface.co/openai/clip-vit-large-patch14}, leveraging both its architecture and pre-trained weights as the foundation for UniSE-CLIP.

\paragraph{UniSE-MLLM.}
The second type is back-ended by \textit{MLLMs (multimodal large language models)} \cite{llava-liu2023visual,wang2024qwen2}, which is more expressive, especially in representing composed inputs. Without losing generality, a composed query is presented by the following template: 
\begin{equation*}
    \text{[Task]: \$task}, \text{[Query]: \$s-tok, \$q-tok, [EOS]}
\end{equation*}
In this place, \$task indicates the task type, \$s-tok stands for the visual tokens from screenshot, while \$q-tok represents the textual tokens from query. The output embedding from the special token [EOS] is used to represent the composed input. For UniSE-MLLM, we adopt Qwen2-VL-2B\footnote{https://huggingface.co/Qwen/Qwen2-VL-2B-Instruct} and initialize it with its pre-trained parameters.

\subsection{Training}
We propose a two-stage training workflow based on the composition of VIRA dataset. First, we perform pre-training using screenshots and their captions, denoted as $\{(s_i,c_i)\}_{N}$. In this stage, the model aims to capture the fine-grained semantic about the screenshot by learning to discriminate its relevant caption from irrelevant ones. Thus, we employ a bidirectional contrastive learning loss, ensuring that the model can correctly match screenshots to their captions and captions to their screenshots:
\begin{equation}
\label{eq-loss1-1}
    \mathcal{L}_{s_1} = \mathcal{L}_{con}(\boldsymbol{e}_s, \boldsymbol{e}_c) + \mathcal{L}_{con}(\boldsymbol{e}_c, \boldsymbol{e}_s)
\end{equation}
where \(\boldsymbol{e}_s\) and \(\boldsymbol{e}_c\) represent the embeddings of screenshots and captions, respectively. The function \(\mathcal{L}_{con}(\boldsymbol{u}, \boldsymbol{v})\) is formulated as:
\begin{equation}
\label{eq-loss1-2}
\mathcal{L}_{con}(\boldsymbol{u}, \boldsymbol{v}) = -\frac{1}{\left |\mathcal{B}\right |} \sum_{i\in \mathcal{B}}^{} \log\frac{\exp(\boldsymbol{u}^{T}_{i}\boldsymbol{v}_{i}/\tau )}{\sum_{j\in \mathcal{B}}^{}\exp(\boldsymbol{u}^{T}_{i}\boldsymbol{v}_{j}/\tau ) }    
\end{equation}
where $\mathcal{B}$ represents the set of in-batch query samples, and $\tau$ is the temperature parameter that controls the strength of penalties on negative samples. 

We continue to fine-tune the pre-trained model based on the question-answering data, denoted as $\{(q_i,s_i)\}_{N}$ ($q_i$ can either be a textual query for a q2s tuple, or a combination of screenshot and its conditioned query for a sq2s triplet.) With this operation, the model is strengthened in handling retrieval-related tasks. In this stage, the following loss function is minimized: 
\begin{equation}
\mathcal{L}_{s_2} = \mathcal{L}_{con}(\boldsymbol{e}_q, \boldsymbol{e}_s)
\label{eq:loss_s2}
\end{equation}
where \(\boldsymbol{e}_q\) and \(\boldsymbol{e}_s\) represent the embeddings of queries and screenshots, respectively. Additional training details and the hyper-parameter settings are provided in~\Cref{appendix:training-details}.



\begin{figure*}[t]
    \centering
    \includegraphics[width=0.95\textwidth]{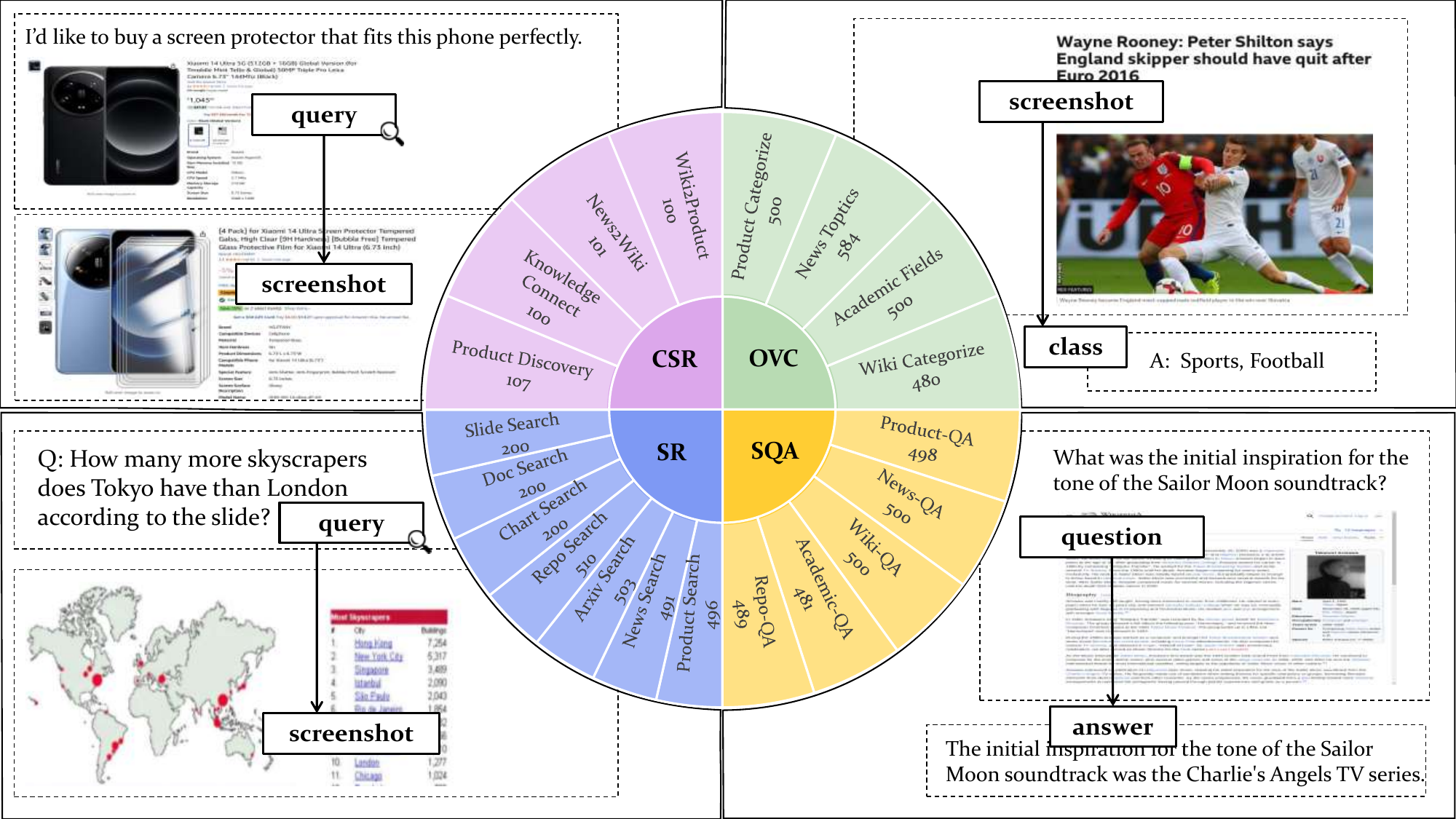}
    \vspace{-0.3cm}
    \caption{MVRB benchmark. There are four task categories: screenshot retrieval, composed screenshot retrieval, screenshot question answering, and open-vocab classification. Each category covers multiple concrete task scenarios.}  
    \vspace{-0.5cm}
    \label{fig:benchmark}
\end{figure*}  


\section{Benchmark: MVRB} 
We create the Massive Vis-IR Benchmark (MVRB), which comprehensively evaluate a retriever's capability in handling general Vis-IR tasks. The outline of MVRB and its creation process is presented below, with more details provided in ~\Cref{appendix:detail-of-MVRB}.

\subsection{Evaluation Tasks} 
Following the practice of popular retrieval benchmarks, such as MTEB \cite{muennighoff2022mteb} and MMEB \cite{jiang2024vlm2vec}, we introduce four task categories, each encompassing multiple tasks relevant to critical application scenarios. 

\subsubsection{Screenshot Retrieval}
Screenshot Retrieval (SR) consists of evaluation samples, each comprising a textual query $q$ and its relevant screenshot $s$: ($q$, $s$). The retrieval model needs to precisely retrieve the relevant screenshot for a testing query from a given corpus $S$. Each evaluation sample is created in two steps: 1) sample a screenshot $s$, 2) prompt the LLM to generate a search query based on the caption of screenshot. We consider seven tasks under this category, including product retrieval, paper retrieval, repo retrieval, news retrieval, chart retrieval, document retrieval, and slide retrieval.

\subsubsection{Composed Screenshot Retrieval}
Composed Screenshot Retrieval (CSR) is made up of sq2s triplets. Given a screenshot $s1$ and a query $q$ conditioned on $s1$, the retrieval model needs to retrieve the relevant screenshot $s2$ from the corpus $S$. 
We define four tasks for this category, including product discovery, news-to-Wiki, knowledge relation, and Wiki-to-product.
All tasks in this category are created by human annotators. For each task, annotators are instructed to identify relevant screenshot pairs and write queries to retrieve \(s_2\) based on \(s_1\). Further details on task definitions and the annotation process are provided in \Cref{appendix:CSR-details}.


\subsubsection{Screenshot Question Answering} 
Screenshot Question Answering (SQA) comprises sq2a triplets. Given a screenshot $s$ and a question $q$ conditioned on $s$, the retrieval model needs to retrieve the correct answer $a$ from a candidate corpus $A$. Each evaluation sample is created in three steps: 1) sample a screenshot $s$, 2) prompt the MLLM to generate a question $q$, 3) prompt the MLLM to generate the answer $a$ for $q$ based on $s$. The following tasks are included in this category: product-QA, news-QA, Wiki-QA, paper-QA, repo-QA. 

\subsubsection{Open-Vocab Classification}
Open-Vocab Classification (OVC) is performed using evaluation samples of screenshots and their textual class labels. Given a screenshot $s$ and the label class $C$, the retrieval model needs to discriminate the correct label $c$ from $C$ based on the embedding similarity. We include the following tasks in this category: product classification, news-topic classification, academic-field classification, knowledge classification. For each task, we employ human labelers to create the label class and assign each screenshot with its correct label. 


\subsection{Optimizations}  
We conduct the following operations to optimize the quality and usability of the benchmark. 

We first introduce \textbf{quality-control} while creating the evaluation samples. Our quality-control framework comprises two main operations: \textbf{automatic assessment} and \textbf{human verification}. We employ multiple MLLMs to constitute a quality-control committee, which make assessment based on three criteria. 1) Clarify, whether the query conveys a concrete information need, 2) Reasonability, whether the query is appropriate in practice, 3) Correctness, whether the query can be addressed by the retrieval target (a screenshot or an answer).  Each evaluation sample is independently reviewed by the MLLMs, and if it fails to meet any criterion, it is removed. The remaining samples are then verified by human labelers using the same principles. An evaluation sample is successfully created only if it passes both quality-control stages. 

We then \textbf{construct the corpus} for each retrieval task. On the one hand, a corpus needs to maintain sufficient difficulty to effectively distinguish the capabilities of different retrievers. On the other hand, its scale should be properly controlled to prevent excessive evaluation costs. Considering these factors, the retrieval corpus is created in two steps.  First, a moderate number of retrieval candidates (around 5,000) are randomly sampled for each task\footnote{This allows users to complete their evaluation in a few hours using one A800-80G GPU, e.g., 3.5h for UniSE-MLLM.}. Second, a group of hard negative samples are introduced for each sample. For (composed) screenshot retrieval, hard negatives are selected based on their similarity to the query. For screenshot question answering, hard negatives are generated by modifying the correct answer using an LLM (see Appendix \ref{appendix:detail-of-MVRB} for details). The hard negatives also undergo quality-control before being added to the corpus. 



\begin{table*}[t]
\footnotesize
\resizebox{\textwidth}{!}
{
\begin{tabular}{@{}cccccccc}
\toprule
\multicolumn{1}{l|}{\multirow{2}{*}{\textbf{Models}}} & \multicolumn{1}{c}{\multirow{2}{*}{\textbf{Backbone}}} & \multicolumn{1}{c|}{\multirow{2}{*}{\textbf{\#params}}} & \multicolumn{4}{c|}{\textbf{Per Meta-Task Score}}                                                        & \multicolumn{1}{c}{\multirow{2}{*}{\textbf{Overall}}} \\ \cmidrule(lr){4-7}
\multicolumn{1}{l|}{}                                 & \multicolumn{1}{c}{}                                   & \multicolumn{1}{c|}{}                                   & \multicolumn{1}{c}{SR} & \multicolumn{1}{c}{CSR} & \multicolumn{1}{c}{SQA} & \multicolumn{1}{c|}{OVC}    & \multicolumn{1}{c}{}                              \\ \midrule
\rowcolor{gray!10} 
\multicolumn{1}{l|}{\texttt{number of datasets}}    &  -- &\multicolumn{1}{c|}{--}       & \texttt{7}             & \texttt{4}   & \texttt{5}   & \multicolumn{1}{c|}{\texttt{4}} &\texttt{20}\\\midrule
\multicolumn{8}{c}{\bf{OCR + Text Retrievers}}                                                                                                                                                                                                                                                                                             \\ \midrule
\multicolumn{1}{l|}{BM25 \cite{bm25-robertson2009probabilistic}}                             &            --              & \multicolumn{1}{c|}{--}                                   & 40.71                & 28.51                 & 38.46                  & \multicolumn{1}{c|}{6.23} & 30.81                                                                            \\
\multicolumn{1}{l|}{DPR \cite{dpr}}                              &         BERT-Base                                               & \multicolumn{1}{c|}{109M}                                   & 24.98                 & 18.73                  & 27.35                  & \multicolumn{1}{c|}{22.08} & 23.74                                            \\
\multicolumn{1}{l|}{BGE \cite{bge}}                              &  BERT-Base                                                       & \multicolumn{1}{c|}{109M}                                   & 45.67                 & 36.73                 & 36.29                  & \multicolumn{1}{c|}{47.91} & 41.99                                            \\
\multicolumn{1}{l|}{E5-Mistral \cite{e5-mistral}}                              &  Mistral-7B                                                       & \multicolumn{1}{c|}{7.11B}                                   & 38.68                 & \underline{51.44}                 & 43.00                  & \multicolumn{1}{c|}{45.08} & 45.51                                            \\ \midrule
\multicolumn{8}{c}{\bf{General Multimodal Retrievers}}                                                                                                                                                                                                                                                                                            \\ \midrule
\multicolumn{1}{l|}{VISTA \cite{vista}}                            &   BERT-Base                                                    & \multicolumn{1}{c|}{196M}                                   &    5.21                    &        11.29                 &   25.78                      &   \multicolumn{1}{c|}{16.61}       &          13.85                                         \\
\multicolumn{1}{l|}{Uni-IR \cite{mbeir-wei2023uniir}}                           &   CLIP-Large                                                     & \multicolumn{1}{c|}{428M}                                  & 12.35                 & 35.92                  & 29.68                  & \multicolumn{1}{c|}{20.06} & 19.63                                            \\

\multicolumn{1}{l|}{CLIP \cite{clip-radford2021learning}}                             &                       CLIP-Large                                 & \multicolumn{1}{c|}{428M}                                    & 18.89                 & 25.39                  & 23.90                  & \multicolumn{1}{c|}{30.40} & 23.75                                            \\
\multicolumn{1}{l|}{SigLIP \cite{siglip2023}}                           &                                     SOViT-400m                   & \multicolumn{1}{c|}{878M}                                    & 38.33                 & 34.48                  & 19.60                  & \multicolumn{1}{c|}{40.64} & 33.34                                           \\
\multicolumn{1}{l|}{E5-V \cite{e5-v2024}}                             &  LLaVA-1.6                                                     & \multicolumn{1}{c|}{8.35B}                                    & 34.11                & 26.59                & 5.23                  & \multicolumn{1}{c|}{32.85} & 25.13                                            \\
\multicolumn{1}{l|}{VLM2Vec \cite{jiang2024vlm2vec}}                          &  Phi-3.5-V                                                      & \multicolumn{1}{c|}{4.15B}                                   & 15.93                 & 48.05                 & \textbf{49.42}                  & \multicolumn{1}{c|}{23.24} & 32.19                                           \\
\multicolumn{1}{l|}{MM-Embed \cite{nv-mm-embed2024}}                         &                          LLaVA-1.6                              & \multicolumn{1}{c|}{7.57B }                                   & 25.86                 & 40.93                  & 42.83                  & \multicolumn{1}{c|}{32.67} & 34.48                                \\  \midrule
\multicolumn{8}{c}{\bf{Screenshot Document Retrievers}}                                                                                                                                                                                                                                                                                           \\ \midrule
\multicolumn{1}{l|}{ColPali \cite{faysse2024colpali}}                          &                                  Paligemma              & \multicolumn{1}{c|}{2.92B}                                   & \underline{61.73}                 & 35.00                  & 35.32                  & \multicolumn{1}{c|}{31.04} & 43.64                                            \\ 
\multicolumn{1}{l|}{DSE \cite{ma2024unifying}}                              &                                  Phi-3-V                      & \multicolumn{1}{c|}{4.15B}                                   &61.54                  & 37.78                 &         39.24         & \multicolumn{1}{c|}{31.51} & 45.21                                            \\
\multicolumn{1}{l|}{GME \cite{zhang2024gme}}                              &                                            Qwen2-VL            & \multicolumn{1}{c|}{2.21B}                                  & {61.62}                 & 37.68                  & 37.78                  & \multicolumn{1}{c|}{\underline{47.98}} & \underline{48.14}                                            \\ 
\multicolumn{1}{l|}{\textbf{UniSE-CLIP (ours)}}                       &                                                 CLIP-Large       & \multicolumn{1}{c|}{428M}                                   &      35.95 &            43.38              &               28.13           & \multicolumn{1}{c|}{40.62}       &     36.41                                             \\
\multicolumn{1}{l|}{\textbf{UniSE-MLLM (ours)}}             &                                                Qwen2-VL        & \multicolumn{1}{c|}{2.21B}           & \textbf{69.63}                 & \textbf{54.49}                & \underline{43.20}                  & \multicolumn{1}{c|}{\textbf{48.26}} & \textbf{55.72}                                            \\ \bottomrule
\end{tabular}
}
\vspace{-8pt}
\caption{Overall performance on MVRB (measure by Recall@1). The aggregation result for each task category and the average score is reported for each method. Top scores are marked in \textbf{bold} while runner-ups are \underline{underlined}.} 
\vspace{-15pt}
\label{tab:main results on MVRB}
\end{table*}



\section{Experiments} 
Our experiments focus on three main perspectives. \textbf{1}. A comprehensive study for existing methods' abilities in Vis-IR. \textbf{2}. The value brought by VIRA dataset. \textbf{3}. The improvement achieved by UniSE. 


\subsection{Baselines}

Our baseline methods fall into three main categories. (1) \textit{OCR + Text Retrievers}, which leverage advanced OCR tools\footnote{https://github.com/PaddlePaddle/PaddleOCR.} to extract text from screenshots, followed by textual retrieval. (2) \textit{General Multimodal Retrievers}, which are vision-language embedding models fine-tuned on diverse multimodal retrieval tasks, such as image retrieval, composed image retrieval, and visual question answering. In our experiments, these models process screenshots in a zero-shot manner. (3) \textit{Screenshot Document Retrievers}. These models are built upon MLLMs architectures and fine-tuned by screenshot-based document retrieval data. Although these models are trained on datasets involving screenshots, their effectiveness is severely constrained by limited data and task diversity, making them insufficient for general Vis-IR tasks. 
We adopt multiple popular methods for each category to ensure a thorough comparison. Detailed specifications about the baseline methods are provided in Appendix~\ref{appendix-baselines}. 


\subsection{Main Results}
The overall performance of baseline retrievers and our UniSE models are presented in Table~\ref{tab:main results on MVRB}. We report the aggregation result for each task category, leaving the detailed performance for each task presented in Appendix~\ref{appendix-detailed results}. We have identified the following key observations from this table:

\noindent\textbf{(1) Our UniSE models achieve the leading performance across all task categories on MVRB.} Specifically, UniSE-MLLM surpasses the previous best screenshot document retriever, GME~\cite{zhang2024gme}, by 7.6\% in average score. Additionally, the UniSE-CLIP model, with only 428 million parameters, outperforms all baselines of no greater sizes while delivering comparable performance to MM-Embed~\cite{nv-mm-embed2024}, the strongest general multimodal retriever, which has 7.57 billion parameters. The superiority of UniSE models verifies the effectiveness of our VIRA dataset. 

\noindent\textbf{(2) The Vis-IR paradigm offers significant advantages over traditional text-based retrieval.}. As shown in~\Cref{tab:main results on MVRB}, even with a complex retrieval workflow powered by advanced OCR tools and powerful text retrievers~\cite{bge}, traditional methods still fall significantly behind UniSE. This suboptimal performance is likely due to the loss of crucial layout and visual semantics, even though OCR can accurately extract textual information. This highlights the great potential of Vis-IR as a unified approach for document retrieval.



\noindent\textbf{(3) Zero-shot application of general multimodal retrievers results in suboptimal performance for Vis-IR}. While these methods can be directly applied to screenshot data, most of them perform worse than other baselines. For instance, the average score of MM-Embed and MLV2Vec falls significantly behind that of GME and DSE, despite using similar MLLM backbones and being fine-tuned on massive datasets. This highlights a key distinction between screenshots and common multimodal data, as the latter lacks a rich combination of data elements presented in visual forms. 

\noindent\textbf{(4) Existing screenshot document retrievers cannot comprehensively address Vis-IR tasks.} Although baselines like ColPali, DSE, and GME achieve substantial improvements in average score, they still underperform other methods in CSR and SQA tasks. Additionally, these methods exhibit inconsistent performance across different domains, as shown in Table \ref{tap:result-of-domains}. In contrast, UniSE-MLLM maintains the leading performance across both task categories and application domains, which demonstrates its well-rounded Vis-IR capabilities. 




\begin{table}[t]
\centering
\begin{adjustbox}{width=0.475\textwidth}
\begin{tabular}{@{}l|cccccc}
\toprule
\multirow{2}{*}{\textbf{Models}} & \multicolumn{6}{c}{\textbf{Domains}} \\ \cmidrule(l){2-7} 
 & Prod. & News & Wiki & Paper & Repo & Others \\ \midrule
\rowcolor{gray!10} 
\multicolumn{1}{l|}{\texttt{\#Datasets}}  & \texttt{4}    & \texttt{3}   & \texttt{3}   & {\texttt{3}} &\texttt{2} &\texttt{5}\\\midrule
BM25 & 28.65 & 28.23 & 21.36 & 39.72 & 45.67 & 34.65 \\
DPR & 17.28 & 33.64 & 25.96 & 21.47 & 33.08 & 27.07 \\
BGE & 38.08 & 48.44 & 34.77 & 41.11 & 51.81 & 47.66 \\ \midrule
VISTA & 14.49 & 19.76 & 17.48 & 11.53 & 20.39 & 6.39 \\
Uni-IR & 19.34 & 24.26 & 23.62 & 8.87 & 18.54 & 21.56 \\
CLIP & 25.96 & 35.57 & 27.70 & 12.14 & 21.75 & 20.27 \\
SIGLIP & \underline{41.46} & 38.73 & 30.19 & 21.09 & 36.10 & 31.76 \\
E5-V & 18.09 & 28.28 & 20.81 & 21.32 & 21.11 & 35.37 \\
VLM2Vec & 28.41 & 38.56 & \underline{39.60} & 20.74 & 37.97 & 31.50 \\
MM-Embed & 31.72 & 40.33 & 36.84 & 23.52 & 42.29 & 35.22 \\ \midrule
ColPali & 36.24 & 41.83 & 26.38 & 47.96 & 55.28 & \underline{53.76} \\
DSE & 37.05 & 48.60 & 28.52 & 46.83 & 60.47 & 52.23 \\
GME & 34.73 & \underline{56.53} & 37.17 & \underline{50.08} & \underline{62.64} & 53.46 \\
UniSE-CLIP & 37.92 & 43.37 & 36.10 & 26.45 & 38.28 & 36.44 \\
Uni-MLLM & \textbf{49.33} & \textbf{64.30} & \textbf{44.50} & \textbf{54.12} & \textbf{70.17} & \textbf{57.61} \\ \bottomrule
\end{tabular}
\end{adjustbox}
\vspace{-5pt}
\caption{Average performance (SR, CSR, SQA, OVC) on different domains. 
Recall@1 is the evaluation metric.} 
\vspace{-10pt}
\label{tap:result-of-domains}
\end{table}

\subsection{Data Ablation} 
We perform extensive ablation studies to analyze the value from VIRA dataset.  First, we examine the impact from captions and question-answering data, which are used for pre-training and fine-tuning, respectively. Next, we look into the roles of different question-answering data, including the s2q tuples and sq2s triplets. Finally, we assess the performance gain from using hard negatives. 

\subsubsection{Two Annotations}
\label{subsubsection-training-data} 
We first explore the impact from the two types of annotations within our VIRA dataset, as shown in Table \ref{tab:data-ablation}. We compare three approaches: the first one relies solely on screenshot captions (first row), the second one leverages only question-answering data (second-row), the last one uses both captions and question-answering data (last row). To ensure a fair comparison, these methods are fine-tuned for the same number of training steps.

We derive the following key observations from the experiment results. First, UniSE demonstrates strong Vis-IR capabilities solely with pre-training on screenshot-caption data, already surpassing ColPaLI~\cite{faysse2024colpali} by 2.5\% in overall score. While the alignment between screenshots and captions differs significantly from downstream tasks, it enables the model to capture fine-grained screenshot semantics, laying a solid foundation for further training. Second, question-answering data alone yields even stronger performance, surpassing the caption-only method by 7.8\%. Third, the combined use of both captions and question-answering data provides additional improvements, achieving performance gains of 9.6\% and 1.8\% over methods using only a single type of annotation data. 

\subsubsection{Composite QA Data} 
We further investigate the benefits of using composite question-answering data, specifically q2s tuples and sq2q triplets, as shown in Table \ref{tab:data-ablation}. To assess their individual contributions, we conduct two ablation studies: one using only q2s tuples (3rd row, w/o sq2q) and the other using only sq2q triplets (4th row, w/o q2s). The results indicate that both q2s tuples and sq2q triplets significantly enhance UniSE’s overall performance, improving the average score by 6.8\% and 8.6\%, respectively, compared to the screenshot-caption pre-trained model. Besides, sq2q triplets provide an extra 1.8\% gain over q2s tuples alone. Finally, the effects of both data types are complementary, as their joint usage leads to the highest performance (6th row, use all). 

\subsubsection{Domain Diversity} 
We examine the detailed impact of using diverse data from multiple domains. To this end, we introduce an ablation approach that includes question-answering data solely from Wikipedia while retaining diverse caption data (5th row, w/o. diversity). The experimental results indicate a clear advantage of using diverse training data, as the default method (6th row, use all) outperforms the ablation approach by 2.2\% in overall performance. This underscores the value of domain diversity in providing uniform support across various Vis-IR tasks. 


\subsubsection{Hard Negatives}
\label{subsubsection-impact-hn}
Finally, we explore the impact of incorporating hard negatives into the training data. In our experiment, we use three ablation methods, as shown in Table \ref{tab:impact-hard-negative}: 1) without hard negatives (1st row), 2) with only sq2s hard negatives (2nd row), and 3) with only s2q hard negatives (3rd row). Compared to the method without hard negatives, introducing either type of hard negatives leads to improved overall performance. Furthermore, combining both types of hard negatives results in the best performance. While the quality of hard negatives could be further improved with more costly processing, the current results offer valuable support for enhancing Vis-IR models developed from the VIRA dataset.

\begin{table}[t]
\small
\centering
\begin{adjustbox}{width=0.475\textwidth}
\begin{tabular}{@{}c|cc|c|c@{}}
\toprule
\multicolumn{1}{c|}{\multirow{2}{*}{\textbf{Caption}}} & \multicolumn{3}{c|}{\textbf{Instructions}}                                                              & \multirow{2}{*}{\textbf{Overall Score}} 
\\ \cmidrule(lr){2-4}
\multicolumn{1}{c|}{}                                  & \text{\textbf{q2s}}         & \text{\textbf{sq2s}}        & \multicolumn{1}{c|}{\textbf{Diversity}} &                                         
\\ 
\midrule
\checkmark                          & \multicolumn{1}{c}{\crossmark} & \multicolumn{1}{c|}{\crossmark} & --                  & {46.14}                    

\\
\crossmark                                  & \checkmark                     & \checkmark                     & \checkmark     &    53.95                                     \\ \midrule
\checkmark                                       & \checkmark                     & \crossmark                     & \checkmark     &      52.94                                   \\
\checkmark                                     & \crossmark                     & \checkmark                     & \checkmark     &      \underline{54.72}                                   \\
\checkmark                                          & \multicolumn{1}{c}{\checkmark} & \multicolumn{1}{c|}{\checkmark} & \crossmark    & {53.53}                    \\ \midrule
\checkmark                                        & \checkmark                     & \checkmark                     & \checkmark     &  \textbf{55.72}                                        \\ \bottomrule
\end{tabular}
\end{adjustbox}
\vspace{-5pt}
\caption{UniSE-MLLM's performance from different data, including 1. caption-only, 2. question-answer only, i.e., q2s and sq2s, 3. w/o. sq2s, 4. w/o. q2s, 5. w/o. diversity, using Wiki as the only source, 6. using all. }
\label{tab:data-ablation}
\end{table}

\begin{table}[t]
\small
\centering
\begin{adjustbox}{width=0.475\textwidth}
\begin{tabular}{cc|cc@{}}
\toprule
\multicolumn{2}{c|}{\textbf{Hard Negatives}}                   & \multicolumn{2}{c}{\textbf{Overall Score}}  \\ \cmidrule{1-4}
\centering \text{\textbf{q2s}}     &   \centering  \text{\textbf{sq2s}}  &\text{\textbf{UniSE-MLLM}} &\textbf{UniSE-CLIP}   \\ \midrule
\crossmark & \crossmark        &        54.26 & 34.99 \\ 
\crossmark & \checkmark               &  \underline{54.68} &   \underline{36.19}                             \\
\checkmark & \crossmark     &      54.50                   &   36.08 \\
\checkmark & \checkmark       &   \textbf{55.72}        &  \textbf{36.41} \\ \bottomrule
\end{tabular}
\end{adjustbox}
\caption{Impact from different hard negative configurations. \text{\textbf{q2s}}: incorporate hard negatives for \text{q2s} tuples; \text{\textbf{sq2s}}: incorporate hard negatives for \text{sq2s} triplets.} 
\vspace{-5pt} 
\label{tab:impact-hard-negative} 
\end{table} 

\section{Related Work}
\paragraph{Neural Document Retrieval.}
Document retrieval is crucial for a wide range of problems, like search engines, open-domain question answering, and retrieval-augmented generation~\cite{dpr,lewis2020retrieval,yuan2023vile}. Traditionally, document retrieval primarily rely on text-based methods, with significant advancements propelled by pre-trained language and large language models. These models have facilitated the development of effective document retrieval systems~\cite{DBLP:conf/emnlp/Ni0LDAMZLHCY22,wang2022text,bge,llama2vec}. However, text-based retrieval methods require prior document parsing and are incapable of handling unstructured information within documents, such as complex layouts, charts, and visual elements~\cite{ma2024unifying}. This limitation results in information loss and highlights the need for new retrieval methods that go beyond text. 


\paragraph{Multimodal Retrieval.} The development of vision-language models (VLMs)~\cite{clip-radford2021learning, align2021, llava-liu2023visual} has significantly advanced the capabilities of general-purpose multimodal retrievers~\cite{mbeir-wei2023uniir,vista,jiang2024vlm2vec,nv-mm-embed2024}, enabling them to perform massive multimodal retrieval tasks, such as text-to-image retrieval~\cite{mscoco-chen2015microsoft}, composed image retrieval~\cite{fashioniq-wu2021fashion,circo}, and visual grounding~\cite{zhu2016visual7w}, etc. However, these methods assume that multimodal documents are well-preprocessed into interleaved image-text sequences, which is unsuitable for the diverse and unstructured nature of many real-world documents. 

Recent studies~\cite{ma2024unifying, faysse2024colpali, zhang2024gme} have explored the use of document screenshots as unified input representations, i.e., visualized information retrieval (Vis-IR). This approach eliminates the need for additional content extraction, preserves all information within the document, and shows potential for effectively handling the complexity of real-world documents. 
Despite these advancements, existing Vis-IR datasets remain limited to specific domains (e.g., Wiki-SS~\cite{ma2024unifying}) and primarily evaluate performance by adapting existing document visual question answering benchmarks to text-to-screenshot retrieval tasks~\cite{slidevqa, mathew2021docvqa, mathew2022infographicvqa, zhu2022towards, li2024multimodal}. This narrow scope restricts the generalizability and applicability of current methods. 
Therefore, there is a clear need for diverse datasets and comprehensive benchmarks to advance this field and enable more robust and scalable solutions.

\vspace{5pt} 
\section{Conclusion} 
In this work, we introduce Vis-IR, a novel paradigm that enables multimodal search in a highly unified manner. Our work makes three fundamental contributions to this emerging field: (1) the creation of VIRA, a large-scale, diverse, and well-annotated dataset for developing Vis-IR models; (2) the development of UniSE, a family of embedding models designed for general Vis-IR applications; and (3) the construction of MVRB, a comprehensive benchmark for evaluating Vis-IR performance. Our experiments uncover the limitations of existing methods in performing Vis-IR tasks and demonstrate the significant improvements brought by VIRA and UniSE. To drive further advancements, we will publicly release all resources and continue expanding in key directions, such as increasing data diversity and incorporating multilingual annotations. 


\clearpage 



\bibliography{custom}

\appendix
\section*{Appendix}
\section{Details of VIRA Dataset Construction}
\label{appendix:details-VIRA}

\subsection{Data Collection}
\label{appedix:VIRA-data-collection}

We collect massive screenshots spanning seven categories. 
Based on the acquisition method, these data primarily fall into two groups: those crawled by us and those curated from publicly available open-source datasets. 
For the crawled data, we use the automated platform \texttt{Playwright}\footnote{https://playwright.dev/python/} to scrape target websites and capture screenshots of the main content pages. To ensure the completeness of the captured images, we perform a verification process using the built-in function of \texttt{Playwright}  and discard any incompletely rendered images. Our entire crawling process adheres to the \texttt{robots.txt} regulations of the respective websites. For the other curated, we have carefully curated the files to ensure they can be converted into complete screenshots, containing both visual and textual content.

\paragraph{Self-Crawled Datasets.} The screenshots obtained through crawling fall into four categories: News, Products, Research Papers, and Project Homepages. The details of the source and processing steps for each category are as follows: 
\begin{itemize}
    \item  \textbf{News}: We scrap news webpages from five mainstream news websites, including BBC, CNN, Fox, CGTN, and Global Times. For each news page, we use \texttt{Playwright} to capture full-page scrolling screenshots and extract text content from the corresponding HTML.
    
    \item \textbf{Products}:  We primarily focus on product detail pages from the Amazon e-commerce website. Utilizing \texttt{Playwright}, we capture screenshots of these pages, which typically feature product images, titles, prices, categories, and other relevant details of the product. Additionally, we extract textual information from the HTML of each product webpage. 
    
    \item \textbf{Research Papers}: We collect research papers from arXiv using the arXiv API\footnote{https://info.arxiv.org/help/api/basics.html} , spanning from January 2018 to November 2024. For each paper, we retain only the most recent version. The original format of these papers is PDF. We convert each page of every paper into an image and extract the text from each page using the \texttt{PyMuPDF}\footnote{https://pypi.org/project/PyMuPDF/} library.
    
    \item \textbf{Project Homepage}: We collect project homepages from GitHub repositories. Initially, we filter out repositories that contain a README.md file. Using \texttt{Playwright}, we capture screenshots of the README region from the homepages of selected repositories and extract the relevant text from corresponding HTML.  Since some README files contain minimal information (e.g., only the repository name), we discard screenshots with a height of less than 300 pixels. 
\end{itemize}

\paragraph{Curated Datasets.} We curate open-source datasets to obtain screenshots for three additional categories: General Documents, Charts, and Common Knowledge. The details of the source and collection process for each category are as follows: 
\begin{itemize}
    \item \textbf{General Documents}: We utilize the publicly available PDFA\footnote{https://hf-mirror.com/datasets/pixparse/pdfa-eng-wds} dataset, a document dataset filtered from SafeDocs. To construct our dataset, we first filter out documents in PDF format and convert each page into an image using \texttt{PyMuPDF}. The source dataset provides OCR data for each document, which we process by stitching together the extracted text from top-left to bottom-right, forming a coherent caption corresponding to each image. 
    
    \item \textbf{Charts}: We utilize the publicly accessible dataset ArxivCap~\cite{li2024multimodal}, which comprises image-caption pairs derived from 572,000 Arxiv papers.
    Each entry contains an image, like a table or figure, and its caption, initially provided as separate text-image pairs rather than a cohesive screenshot. Therefore, we render each image alongside its caption into a single, unified screenshot. 
    
    \item \textbf{Common Knowledge}: We used the publicly available dataset Wiki-SS-Corpus~\cite{ma2024unifying}. This dataset is derived from screenshots of Wikipedia entry webpages and includes caption data for the first 500 words of each page. 
    
\end{itemize}

After collecting the data, we apply a filtering process to ensure content quality and appropriateness. Specifically, we perform keyword matching on the caption text to filter out NSFW content. Additionally, we remove low-quality entries based on aspect ratio and caption length, discarding screenshots with an aspect ratio exceeding 9 and captions with fewer than 100 characters.

\subsection{Data Annotation}
\label{appendix:data-annotation}
\begin{figure*}[tb]
    \centering
    \includegraphics[width=0.95\textwidth]{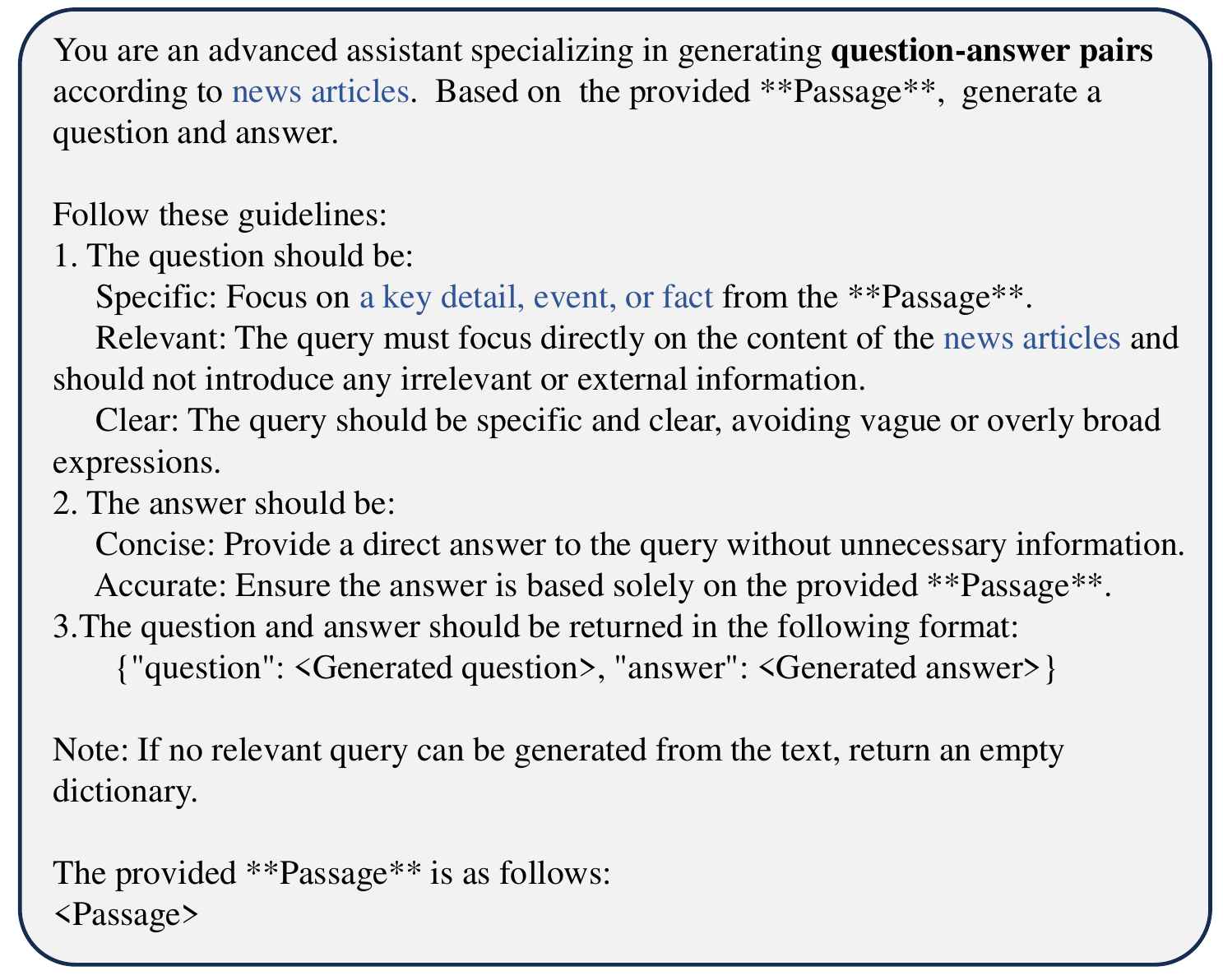}
    \vspace{-0.3cm}
    \caption{The prompt used for q2s annotation. This prompt is designed for the news domain. For other domains, the word in blue can be substituted with the appropriate term for that domain.} 
    \vspace{-0.3cm}
    \label{fig:prompt-q2s}
\end{figure*}

The annotation of VIRA data is categorized into two main types: caption and question-answering.  Caption is the text within the collected screenshots, which is annotated and generated during the screenshot collection process, as detailed in Appendix~\ref{appedix:VIRA-data-collection}. As for the question-answering, we have developed two types of question-answering annotations: q2s tuples and sq2s triplets. The detailed process of creating these question-answering annotations is as follows:

\paragraph{q2s tuples.} A q2s tuple consists of a question \textit{q} and a corresponding screenshot \textit{s} that can be used to answer \textit{q}. Given that we have access to caption for each screenshot, we leverage a large language model (LLM) to generate question-answer pairs closely related to the screenshot. Our approach is similar to the methodology used in constructing Docmatix~\cite{laurenccon2024building}. To enhance annotation quality, we design domain-specific prompts tailored to different categories of screenshots, with the corresponding prompts illustrated in Figure~\ref{fig:prompt-q2s}. For the whole annotation process, we employ the open-source model Qwen2.5-72B-Instruction~\cite{qwen2.5}, which requires approximately 2,078 A800 GPU hours in total.

To enhance the effectiveness of q2s tuples, we augment each tuple with hard negatives. This augmentation process leverages both the text embedding model BGE~\cite{bge} and the visual embedding model EVA-CLIP~\cite{evaclip-sun2023eva}. Specifically, we first encode all captions to construct a corpus embedding and separately encode both the question and the caption of the target screenshot to obtain their respective embeddings. Based on these embeddings, we retrieve the top 15 and top 10 candidates from the corpus for the question and the target screenshot, respectively. To mitigate false negatives, we exclude the top-1 candidate for the question and the top-3 candidates for the target screenshot.

For screenshots in the domains of of News, Products, and Common Knowledge—where natural images frequently appear—we further employ EVA-CLIP to encode the target screenshot and retrieve the top-10 candidates from the corpus, excluding the top-2 to reduce potential false negatives. Finally, we merge all remaining candidate sets and randomly sample 8 candidates to serve as hard negatives for the q2s tuple.

\begin{figure*}[htb]
    \centering
    \includegraphics[width=0.95\textwidth]{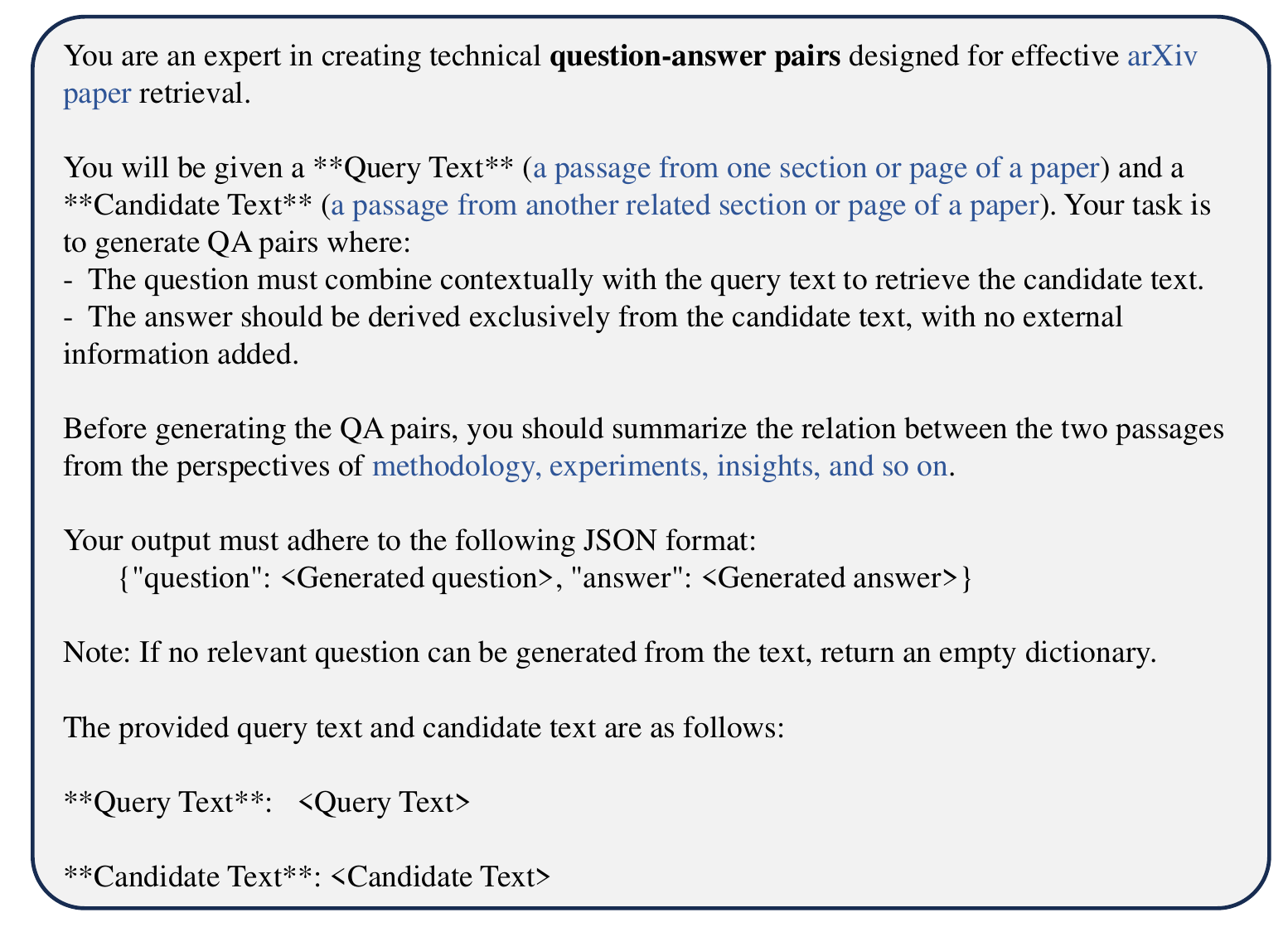}
    \vspace{-0.3cm}
    \caption{The prompt used for sq2s annotation. This prompt is intended for the paper domain. For other domains, the word in blue can be substituted with the appropriate term for that domain.} 
    \vspace{-0.3cm}
    \label{fig:prompt-sq2s}
\end{figure*}

\paragraph{sq2s triplets.} An sq2s triplet consists of two relevant screenshots—a query screenshot and a target screenshot—along with a conditional query.  The target screenshot is used to respond the query conditioned on the query screenshot. To construct an sq2s triplet, we first mine a pair of relevant screenshots from the corpus. Following a similar approach used for augmenting q2s tuples with hard negatives, we employ BGE and EVA-CLIP to retrieve the top-10 candidates and randomly select one as the relevant screenshot for each given screenshot.
For each pair of relevant screenshots, we prompt LLM to analyze their relationship based on their captions and generate a relational query. To enhance annotation quality, we design domain-specific prompts tailored to different categories of screenshot pairs, with the corresponding prompts are illustrated in Figure~\ref{fig:prompt-sq2s} . We utilize the open-source model Qwen2.5-72B-Instruction, and the entire annotation process requires approximately 1002 A800 GPU hours.

Additionally, we augment each sq2s triplet with hard negatives. The conditional query, query screenshot, and target screenshot are used to retrieve candidates from the corpus, following the same methodology applied in augmenting q2s tuples with hard negatives. From the retrieved candidates, 8 candidates are randomly sampled to serve as hard negatives for each sq2s triplet.

\subsection{Statistics of VIRA}
The VIRA is an extensive dataset, comprising a total of 20 million data entries. Of these, approximately 12.96 million are caption data, and around 7.11 million are question-answering data, which includes 5.97 million q2s tuples and 1.14 million sq2s triplets. A detailed breakdown of the data for each type in different domains is provided in Table~\ref{tab:statistic-VIRA}.

\begin{table}[htb]
\centering
\begin{tabular}{ccc}
\toprule
Type & Domain & Number \\
\midrule
\multirow{7}{*}{Caption} & News &  1.85M \\ \cmidrule{2-3}
& Products & 1.24M \\ \cmidrule{2-3}
& Research Papers & 2.39M\\ \cmidrule{2-3}
& Project Homepage & 2.38M \\ \cmidrule{2-3}
& General Documents & 1.82M \\ \cmidrule{2-3}
& Charts & 2.01M\\ \cmidrule{2-3}
& Knowledge & 1.27M\\ 
\hline \hline
\multirow{7}{*}{s2q} & News & 1.24M \\ \cmidrule{2-3}
& Products & 787.5K \\ \cmidrule{2-3}
& Research Papers & 997.4K\\ \cmidrule{2-3}
& Project Homepage & 1.0M \\ \cmidrule{2-3}
& General Documents & 1.09M\\ \cmidrule{2-3}
& Knowledge & 850.2K\\ 
\cmidrule{1-3}
\multirow{8}{*}{qs2s} 
& News & 248.0K \\ \cmidrule{2-3}
& Products & 496.1K\\ \cmidrule{2-3}
& Research Papers & 89.8K\\ \cmidrule{2-3}
& Project Homepage & 69.7K\\ \cmidrule{2-3}
& General Documents & 87.5K\\ \cmidrule{2-3}
& Knowledge & 155.4K\\
\bottomrule
\end{tabular}
\caption{Detailed data counts for each domain in VIRA}
\label{tab:statistic-VIRA}
\end{table}

\section{More Details of UniSE Models}
\subsection{Preprocessing Strategies for Screenshot Images}  
For the UniSE-CLIP model, all input screenshot images are resized to \(224 \times 224\) during both training and evaluating, following the default settings of the original CLIP model. In contrast, the UniSE-MLLM model employs a \textbf{smart resize} strategy for both training and evaluation, which preserves the original aspect ratio of the screenshot images while maximizing their resolution to retain visual quality. Specifically, the maximum image size is set as \(M \times 28 \times 28\), where \(M\) denotes the maximum number of image tokens. For an image with height \(H\) and width \(W\) exceeding this limit, its dimensions are adjusted to \(H' = \lfloor \frac{H}{\beta} \rfloor \times 28\) and \(W' = \lfloor \frac{W}{\beta} \rfloor \times 28\), where \(\beta = \sqrt{ \frac{W \times H}{M \times 28 \times 28}}\). In our UniSE-MLLM, \(M\) is set to 2500.

\subsection{Training Details}
\label{appendix:training-details}

For both UniSE-CLIP and UniSE-MLLM models, training is conducted in two stages: pre-training and instruction fine-tuning. In the pre-training stage, both models are trained on VIRA's screenshot-caption subset, which contains approximately 13 million screenshot-caption pairs. In the instruction fine-tuning stage, the pre-trained models are further refined using VIRA's question-answering subset, comprising approximately 6 million items. For all models and training stages, the initial learning rate is set to \(5 \times 10^{-6}\), with a linear decay schedule applied during training. Other training details are shown below.

\paragraph{UniSE-CLIP Model.}  
In the pre-training stage, the UniSE-CLIP model is trained with a batch size of 8192 for a single epoch. During the instruction fine-tuning stage, the model uses a batch size of 4096 for one epoch, where each query is paired with one positive screenshot and one hard negative sample. All model parameters are updated during both stages of training.

\paragraph{UniSE-MLLM Model.}  
In the pre-training stage, the UniSE-MLLM model is trained with a batch size of 2048 for one epoch. During the instruction fine-tuning stage, the model uses a batch size of 1024 for one epoch, with each query paired with one hard negative sample. Unlike UniSE-CLIP, UniSE-MLLM employs LoRA~\cite{lora} to fine-tune the language model component of Qwen, while all other layers remain frozen throughout the training process. The LoRA rank is set to 32.

\section{Details of MVRB Benchmark Creation}
\label{appendix:detail-of-MVRB}

\begin{figure*}[h]
\centering
    \scriptsize
    \begin{center}
    \tabcolsep 2pt
    \begin{tabular}{m{1.8cm}m{1.8cm}m{3.5cm}m{2.1cm}>{\columncolor{lightgray}}m{3.5cm}}
            \toprule
             \textbf{Task} & \textbf{Subtask} & \textbf{Query Text} & \textbf{Query Image} & \textbf{Target} \\
            \midrule
            \multirow{13}{*}{\makecell[c]{Screenshot\\ Retrieval}}
                  & Product Retrieval
                  & What are the dimensions of the Baosha Women's Small Sling Crossbody bag?
                  & --
                  & \includegraphics[width=15mm,height=15mm]{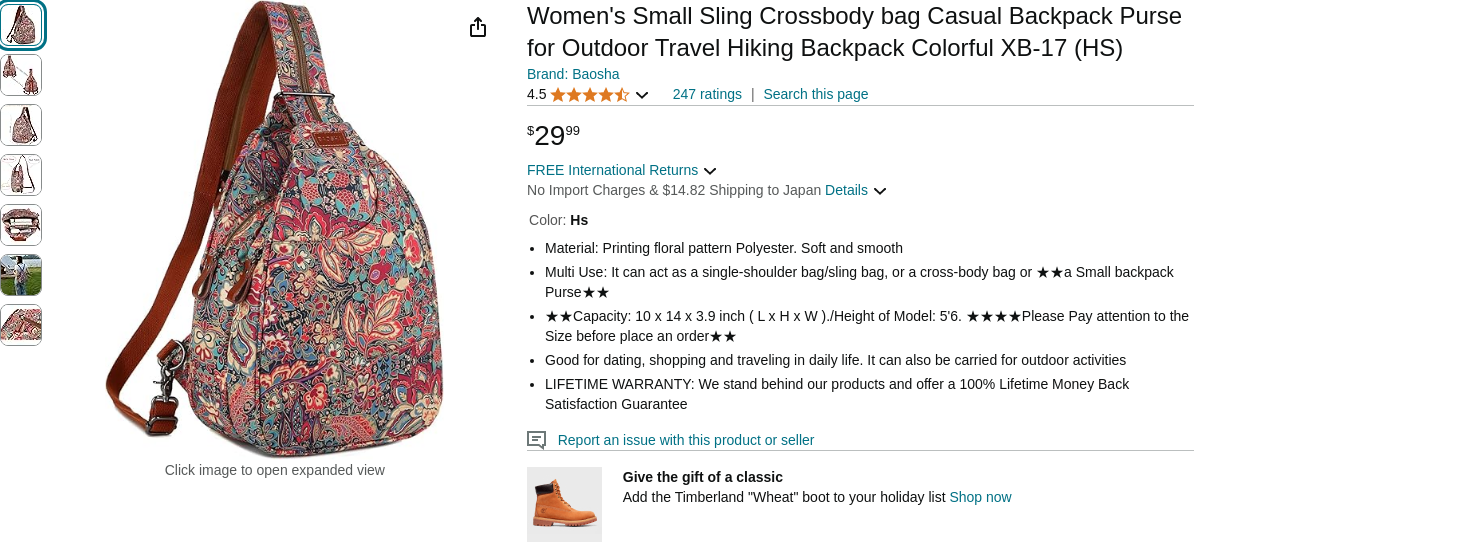}\\
            \cmidrule{2-5}
                  & News Retrieval
                  & What are the four foundational codes that explain Trump's tumultuous path through his life, according to the judge's ruling in the New York fraud case?
                  & --
                  & \includegraphics[width=15mm,height=15mm]{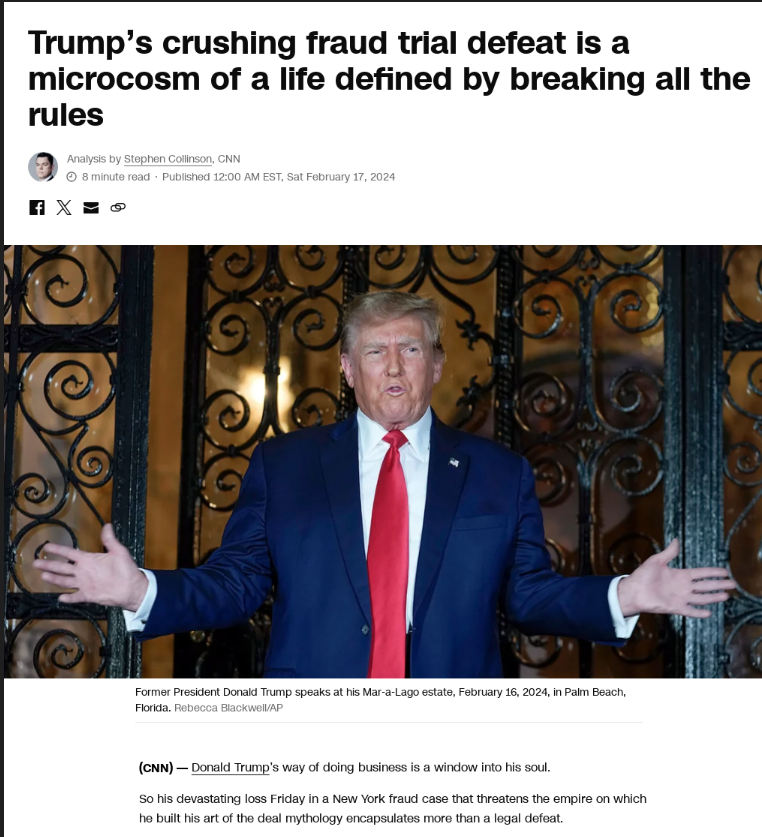}\\
            \cmidrule{2-5}
                  & Paper Retrieval
                  & What is the 'mass twins' scenario in the context of compact stars?
                  & --
                  & \includegraphics[width=15mm,height=15mm]{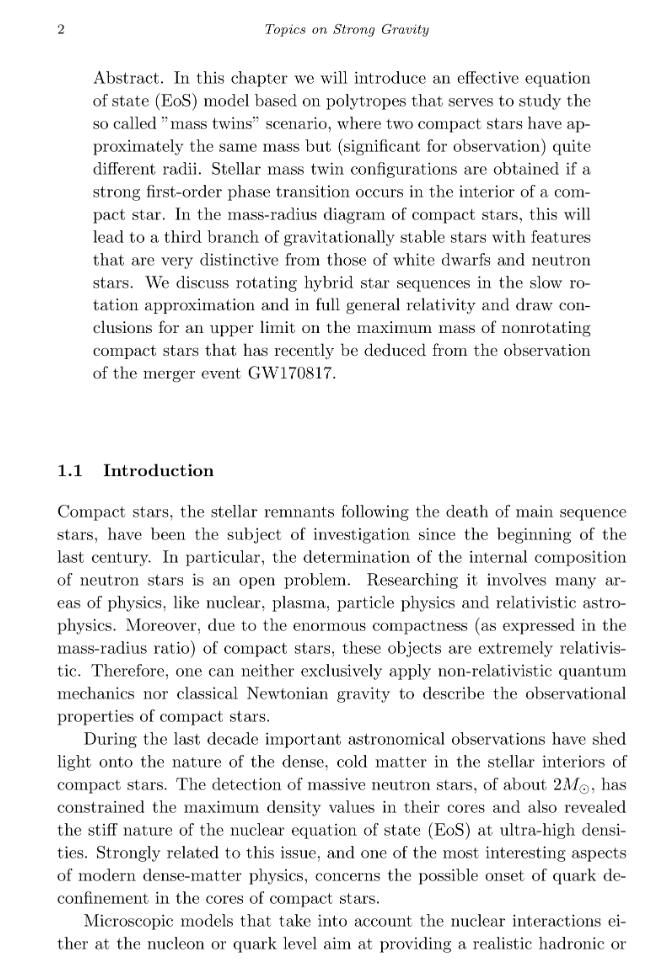}\\
            \cmidrule{2-5}
                  & Chart Retrieval
                  & What percentage of the prize pool did eSports team devils.one receive?
                  & --
                  & \includegraphics[width=15mm,height=15mm]{fig/benchmark_examples/sr4.png}\\
            \cmidrule{2-5}
                  & \makecell[l]{Document\\ Retrieval}
                  & Who is the staff representative mentioned in the document?
                  & --
                  & \includegraphics[width=15mm,height=15mm]{fig/benchmark_examples/sr5.png}\\
            \cmidrule{2-5}
                  & Slide Retrieval
                  & How many more skyscrapers does Tokyo have than London according to the slide?
                  & --
                  & \includegraphics[width=15mm,height=15mm]{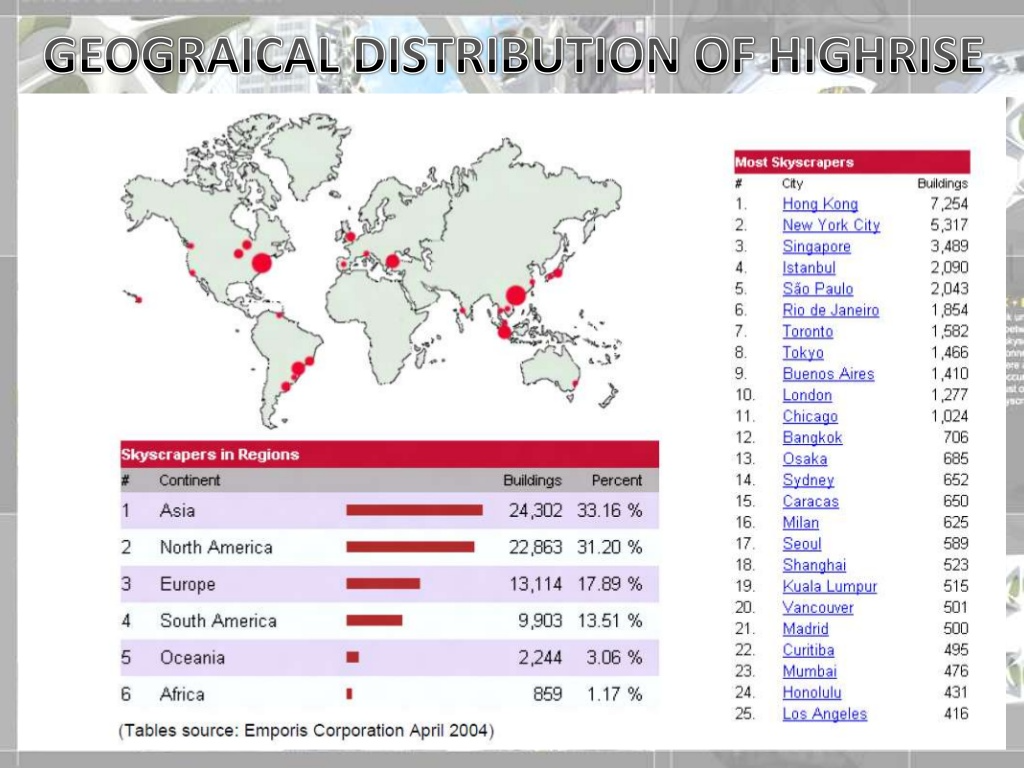}\\
            \cmidrule{2-5}
                  & Repo Retrieval
                  & What are the prerequisites for successfully completing the Knowledge Mining Solution Accelerator?
                  & --
                  & \includegraphics[width=15mm,height=15mm]{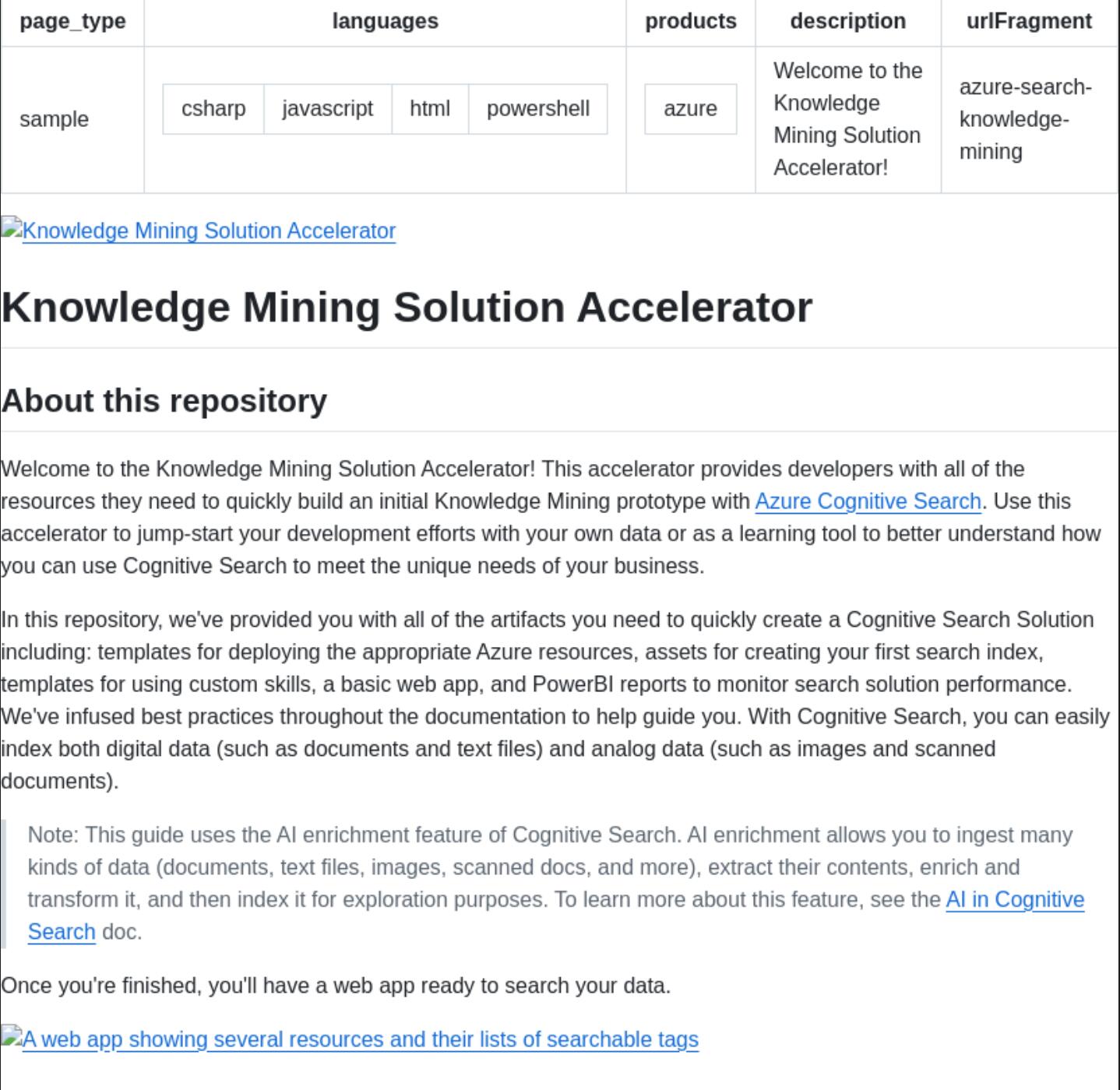}\\
            \midrule
            \multirow{15}{*}{\makecell[c]{Composite\\Screenshot\\ Retrieval}}
                  & \makecell[l]{Product\\ Discovery}
                  & I need a larger size of the same brand.
                  & \includegraphics[width=15mm,height=15mm]{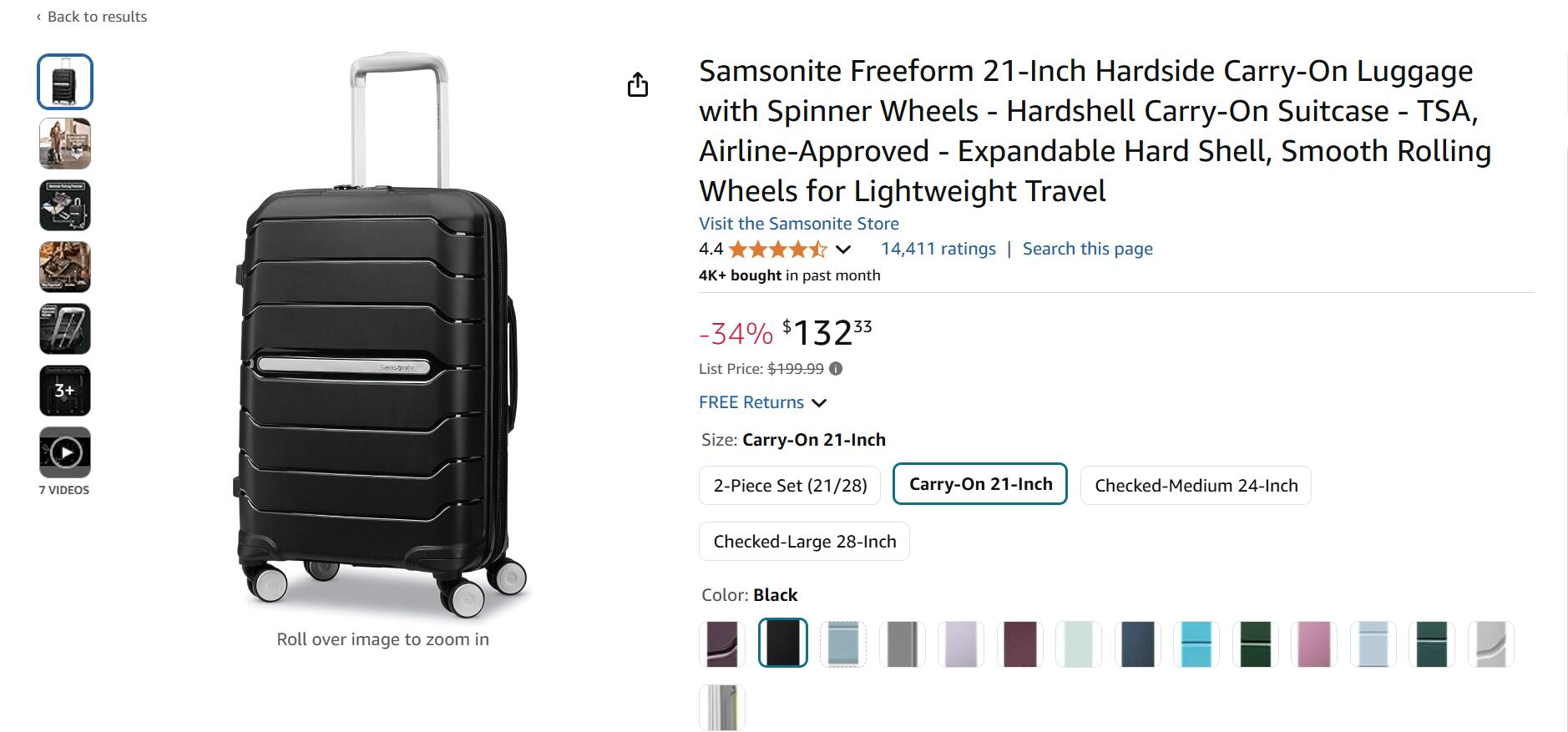}
                  & \includegraphics[width=15mm,height=15mm]{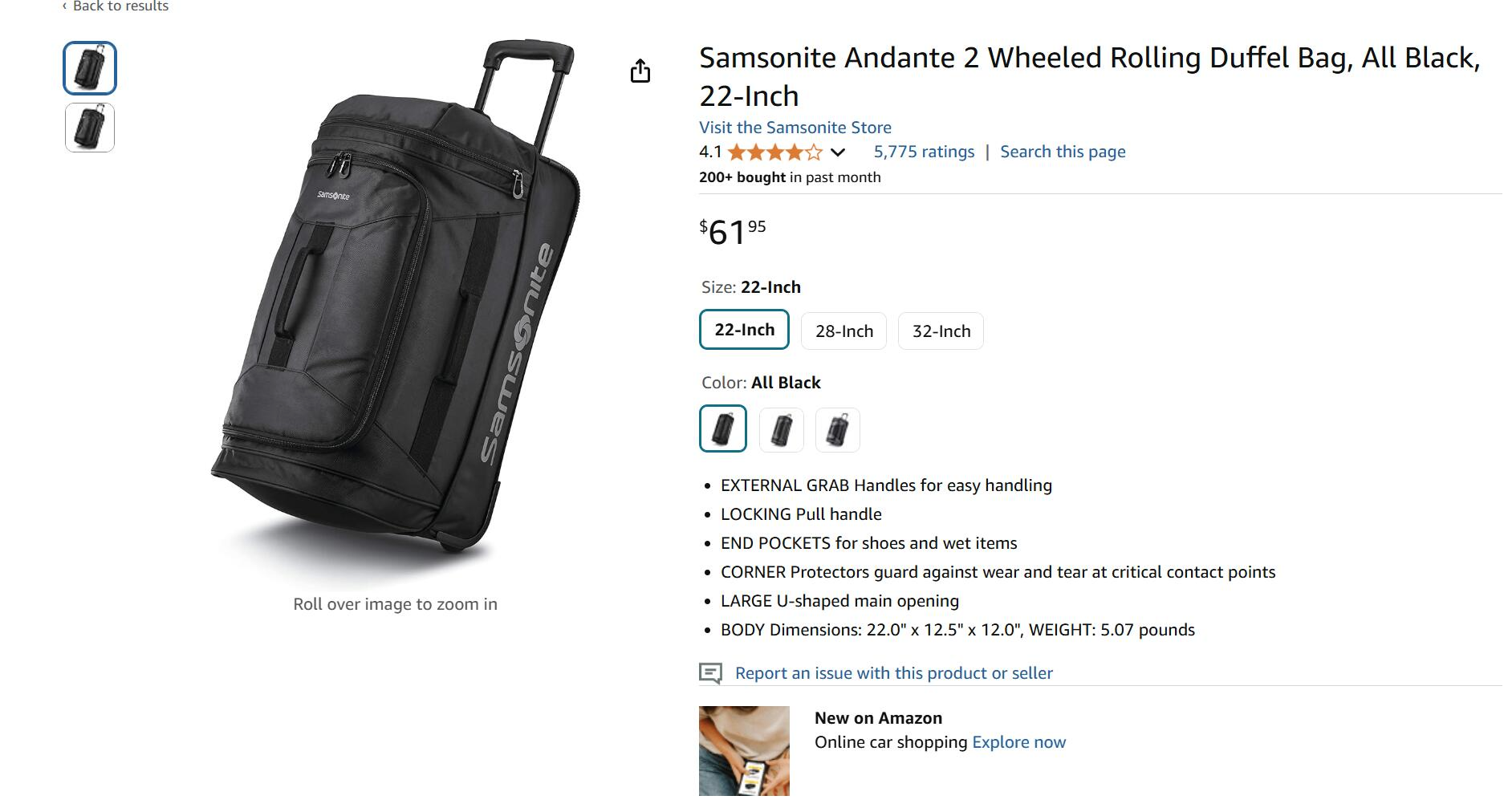}\\
            \cmidrule{2-5}
                  & \makecell[l]{Knowledge\\ Relation}
                  & Which temple is the Buddha statue in the image located?
                  & \includegraphics[width=15mm,height=15mm]{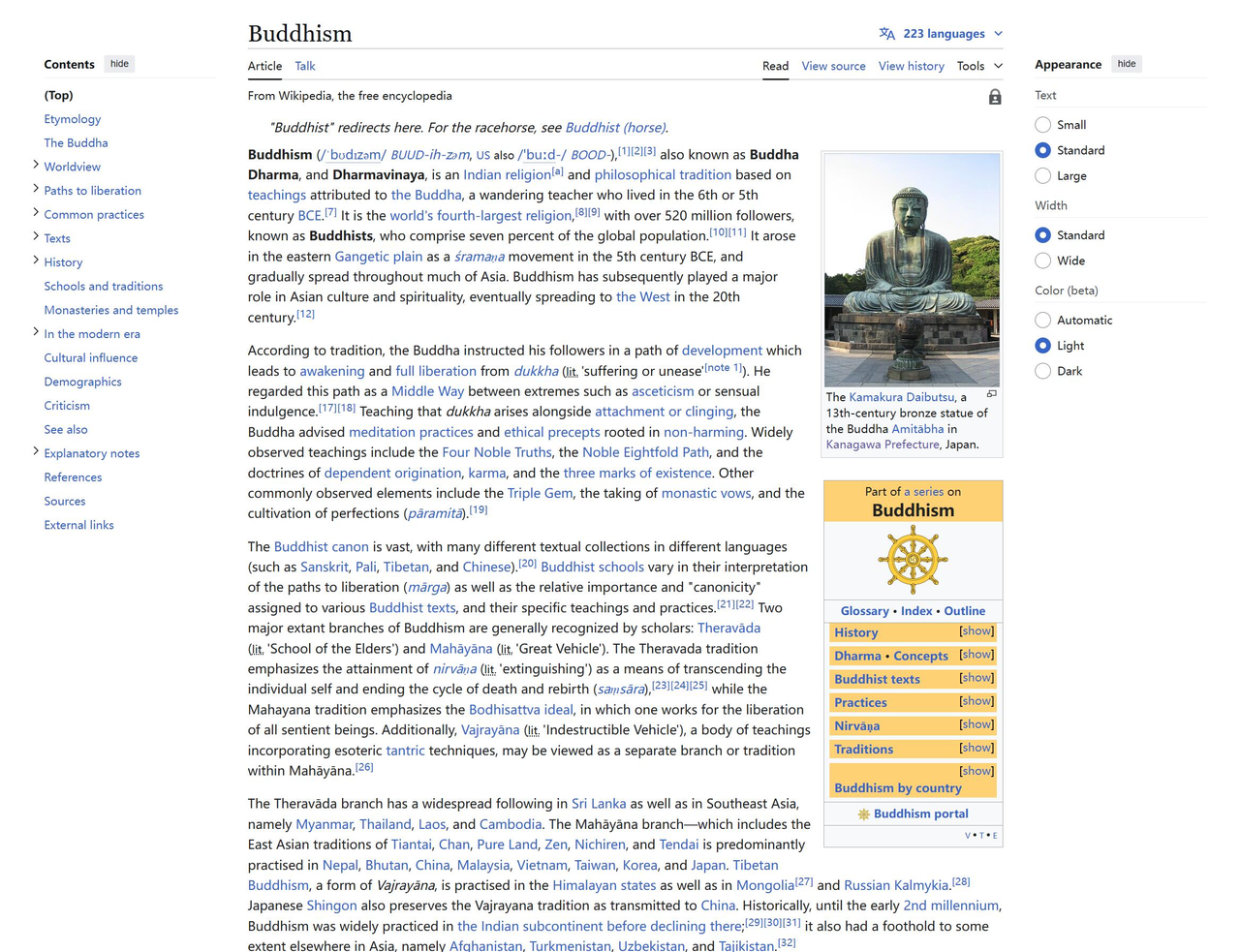}
                  & \includegraphics[width=15mm,height=15mm]{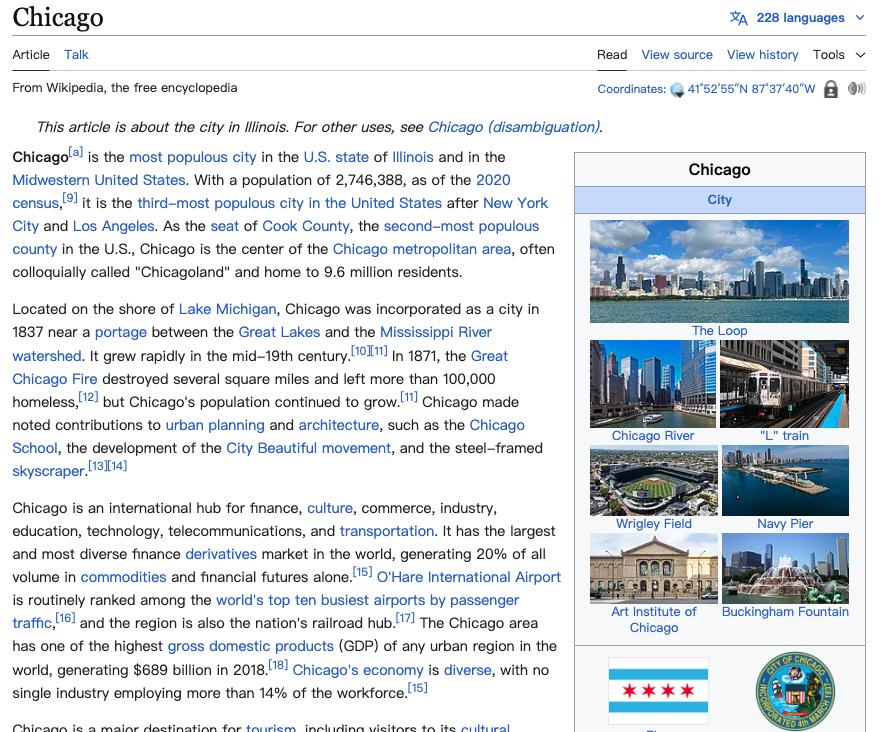}\\
            \cmidrule{2-5}
                  & News-to-Wiki
                  & I need information about the political party of this man in this event.
                  & \includegraphics[width=15mm,height=15mm]{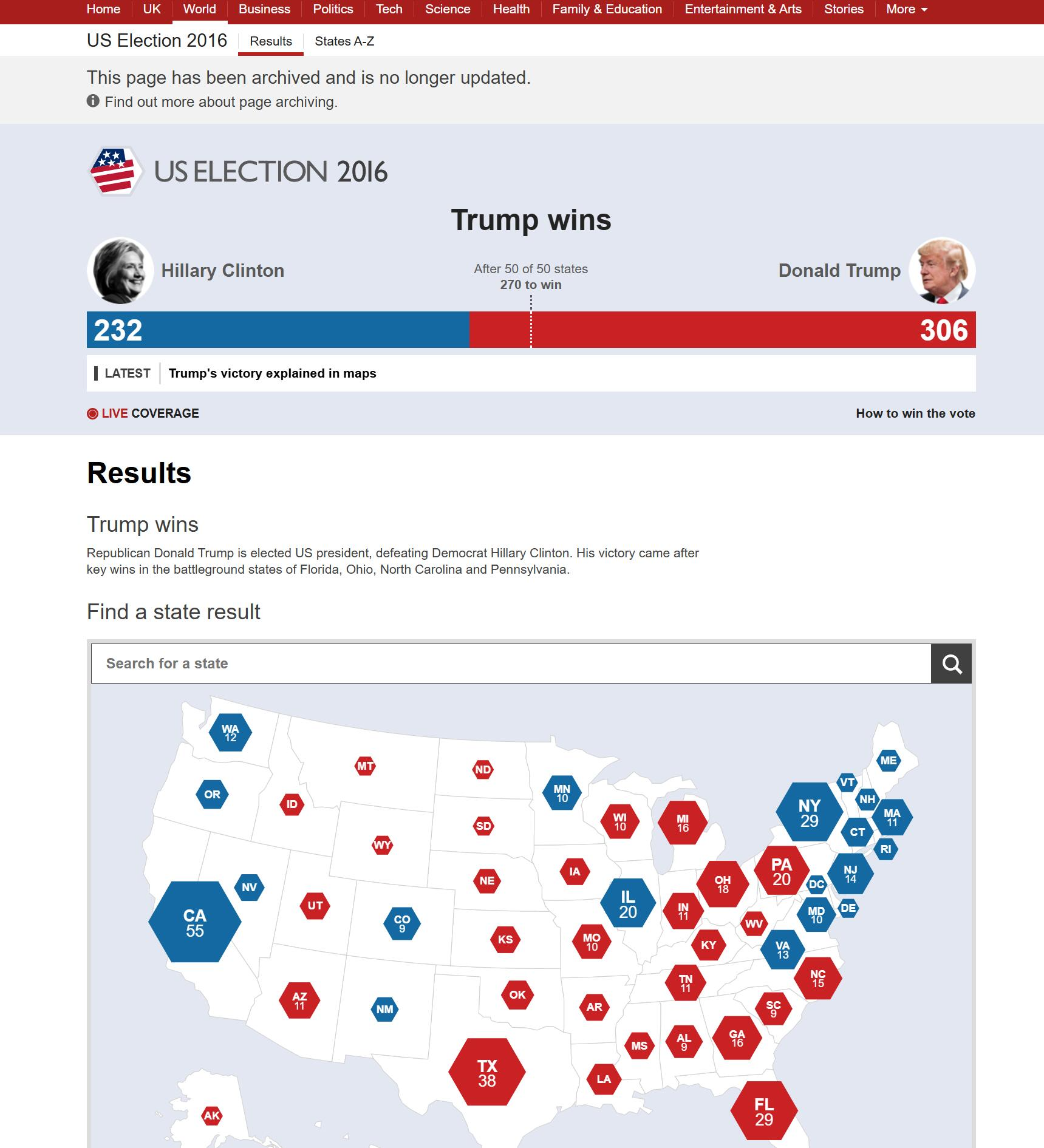}
                  & \includegraphics[width=15mm,height=15mm]{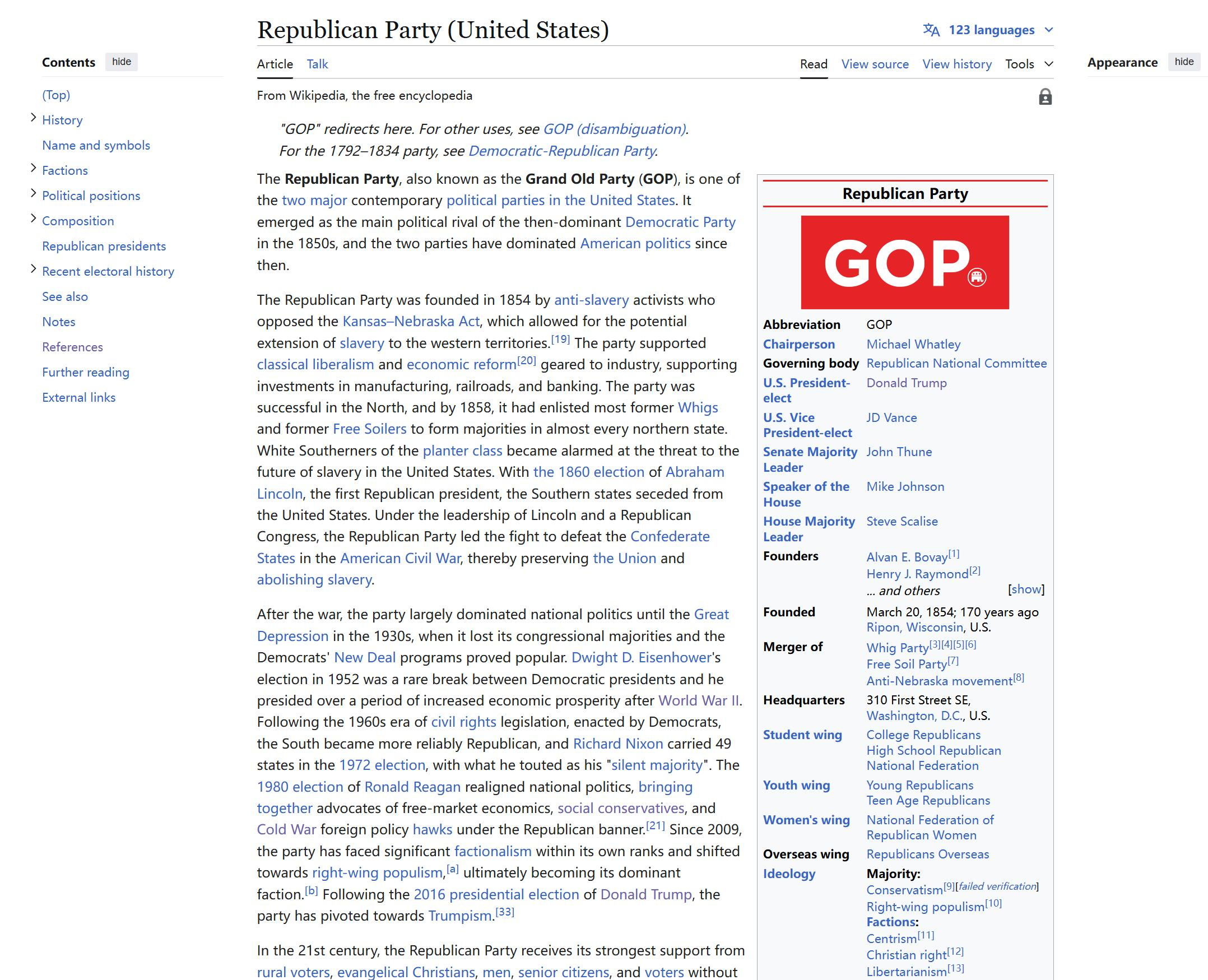}\\
            \cmidrule{2-5}
                  & Wiki-to-Product
                  & Search the most relevant product page.
                  & \includegraphics[width=15mm,height=15mm]{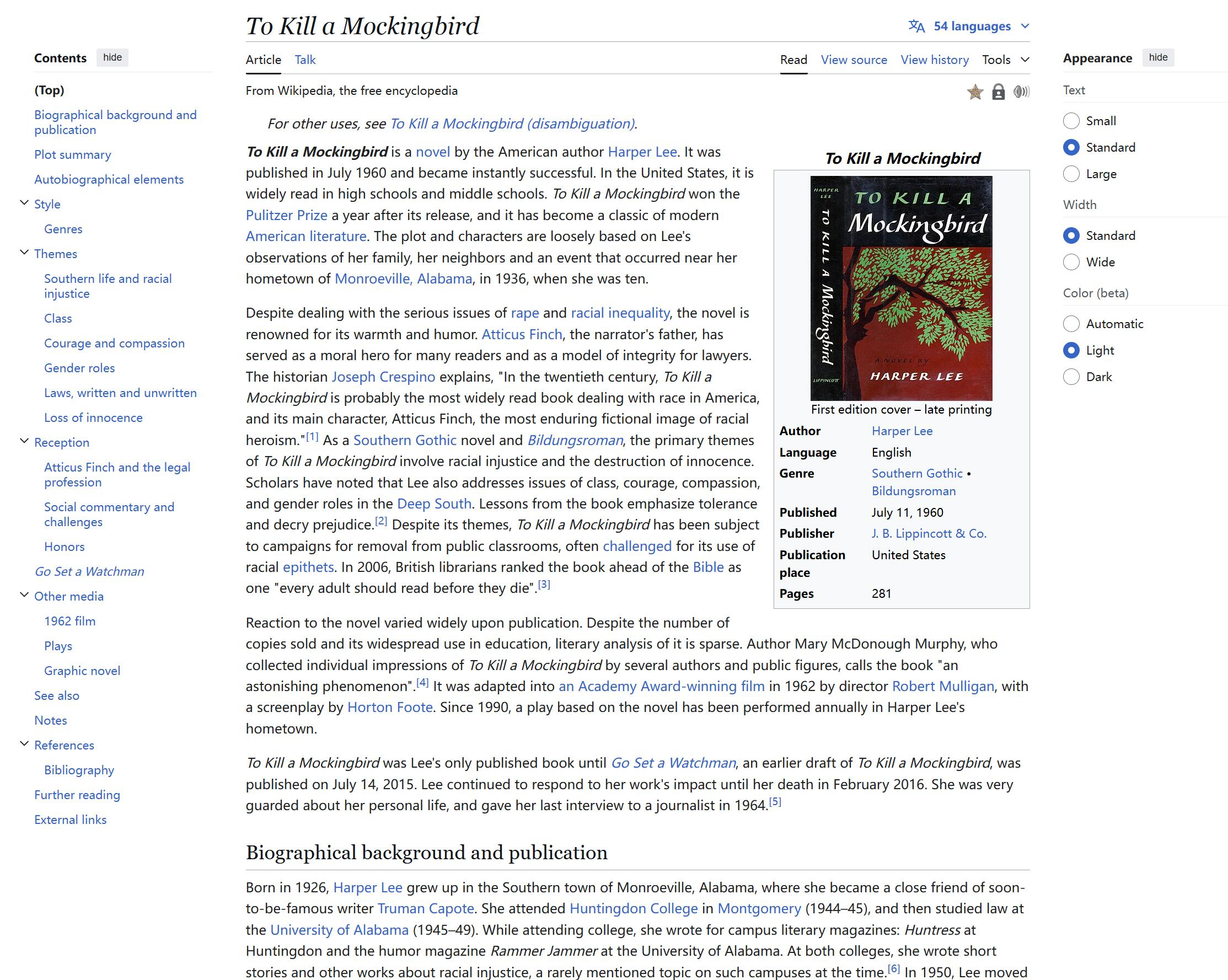}
                  & \includegraphics[width=15mm,height=15mm]{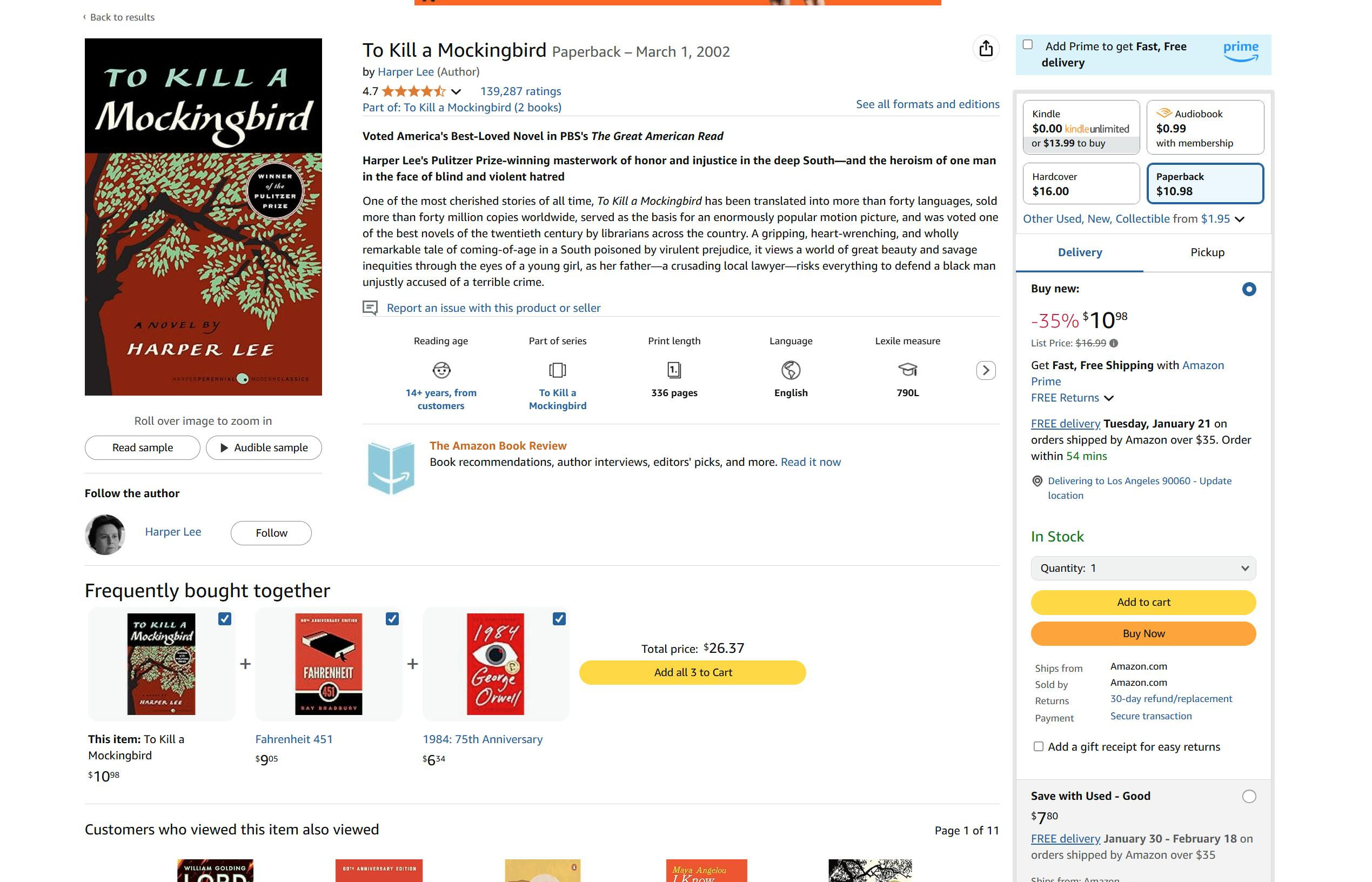}\\
            \bottomrule
    \end{tabular}
    \caption{Examples of tasks in Screenshot Retrieval and Composite Screenshot Retrieval in MVRB.}
    \label{figure:MVRB-example-SR-CSR}
    \end{center}
\end{figure*}

\begin{figure*}[h]
\centering
    \scriptsize
    \begin{center}
    \tabcolsep 2pt
    \begin{tabular}{m{1.8cm}m{1.8cm}m{3.5cm}m{2.1cm}>{\columncolor{lightgray}}m{3.5cm}}
            \toprule
             \textbf{Task} & \textbf{Subtask} & \textbf{Query Text} & \textbf{Query Image} & \textbf{Target} \\
            \midrule
            \multirow{15}{*}{\makecell[c]{Screenshot\\ Question \\ Answering}}
                  & Product-QA
                  & What type of chain is used in the S925 Circle of Life Cremation Memorial Necklace?
                  & \includegraphics[width=15mm,height=15mm]{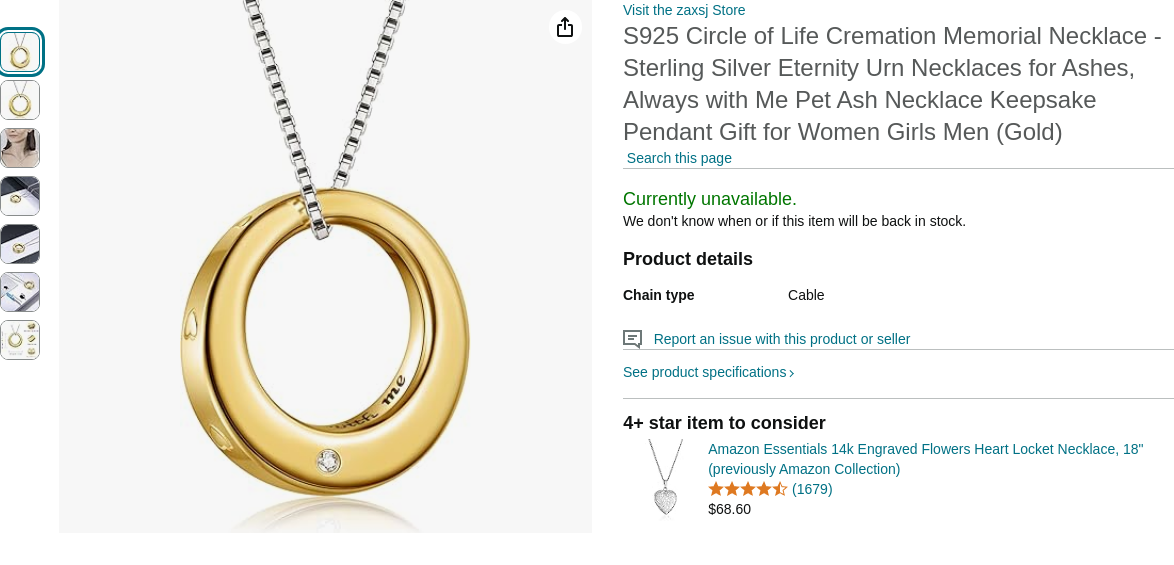}
                  & Cable.\\
            \cmidrule{2-5}
                  & News-QA
                  & What did Jeffrey Garten accidentally send to Ina Garten's publicist?
                  & \includegraphics[width=15mm,height=15mm]{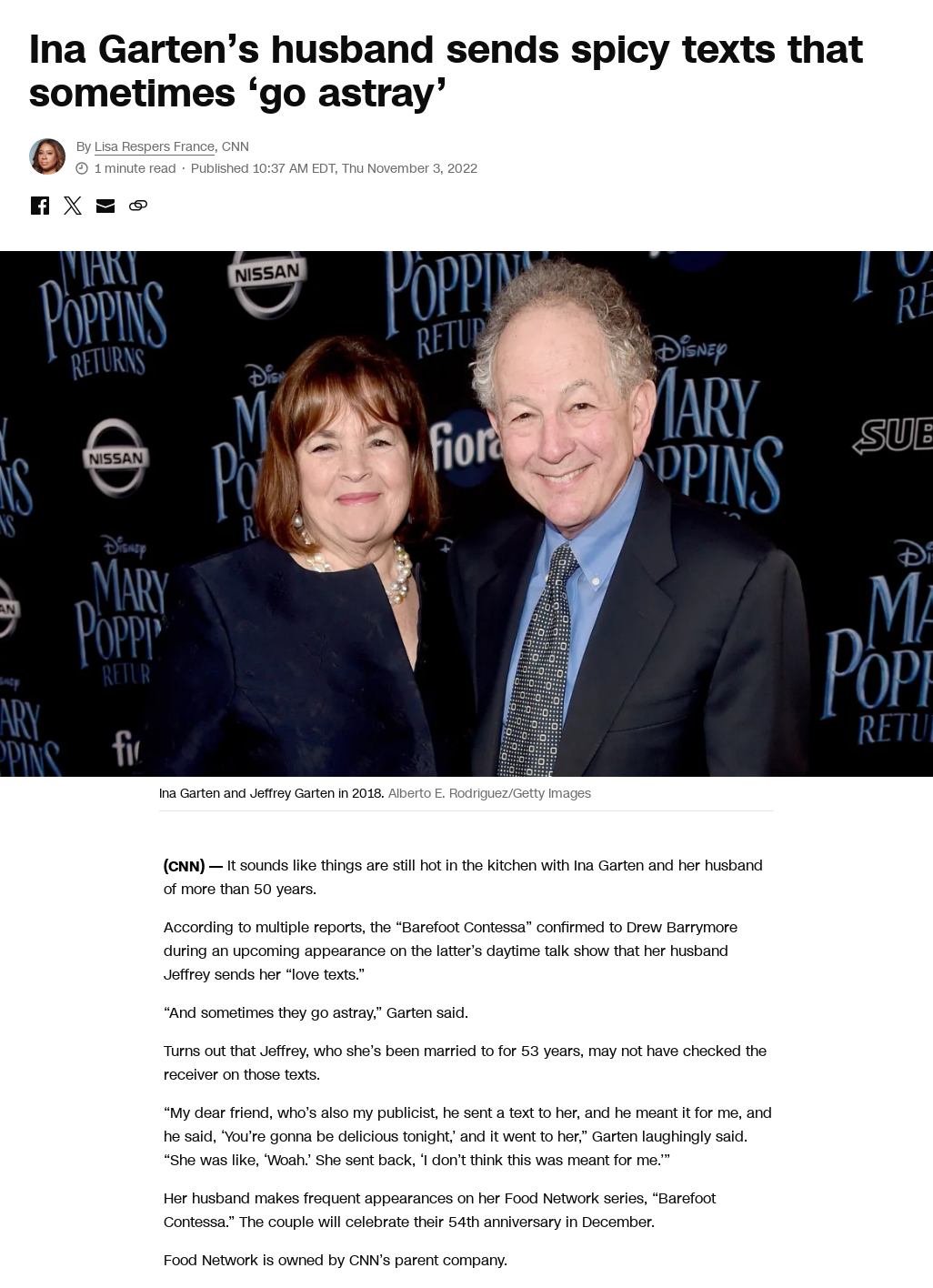}
                  & A text that said, "You're gonna be delicious tonight."\\
            \cmidrule{2-5}
                  & Wiki-QA
                  & What was the initial inspiration for the tone of the Sailor Moon soundtrack?
                  & \includegraphics[width=15mm,height=15mm]{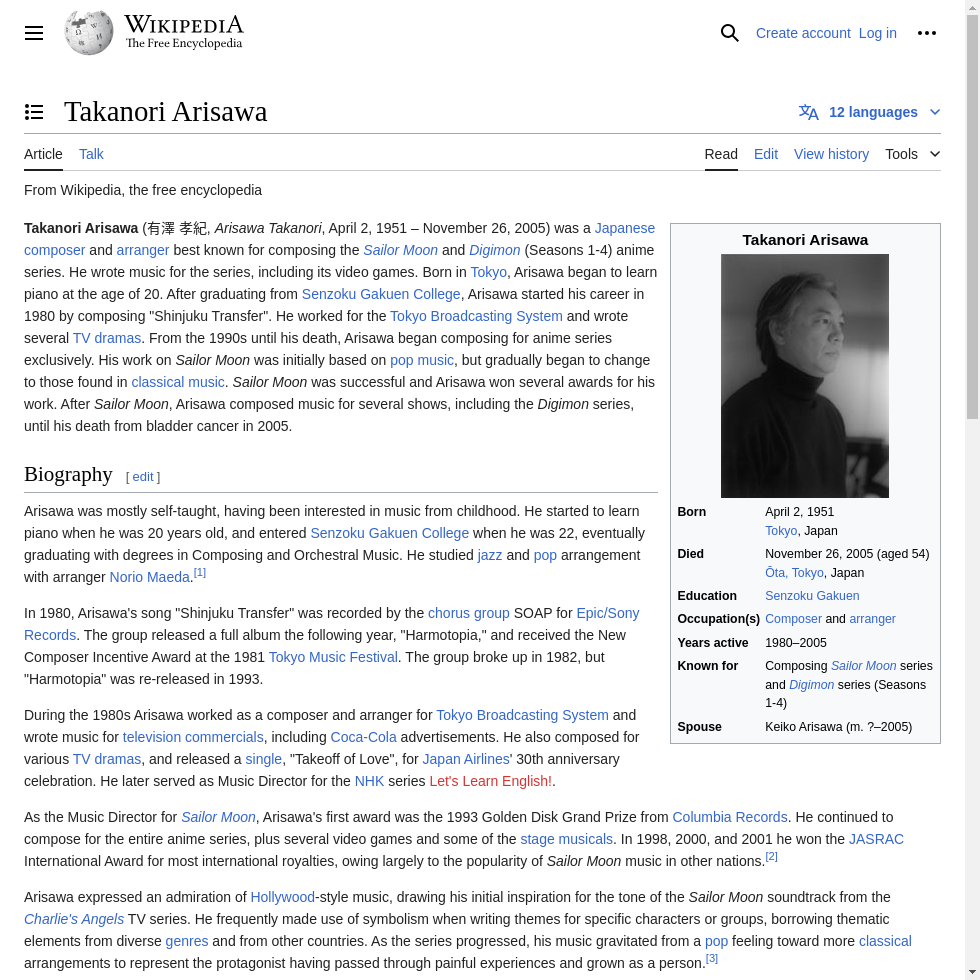}
                  & The initial inspiration for the tone of the Sailor Moon soundtrack was the Charlie's Angels TV series.\\
            \cmidrule{2-5}
                  & Paper-QA
                  & What is the key feature of the optimal contingent debt contract according to Proposition 5?
                  & \includegraphics[width=15mm,height=15mm]{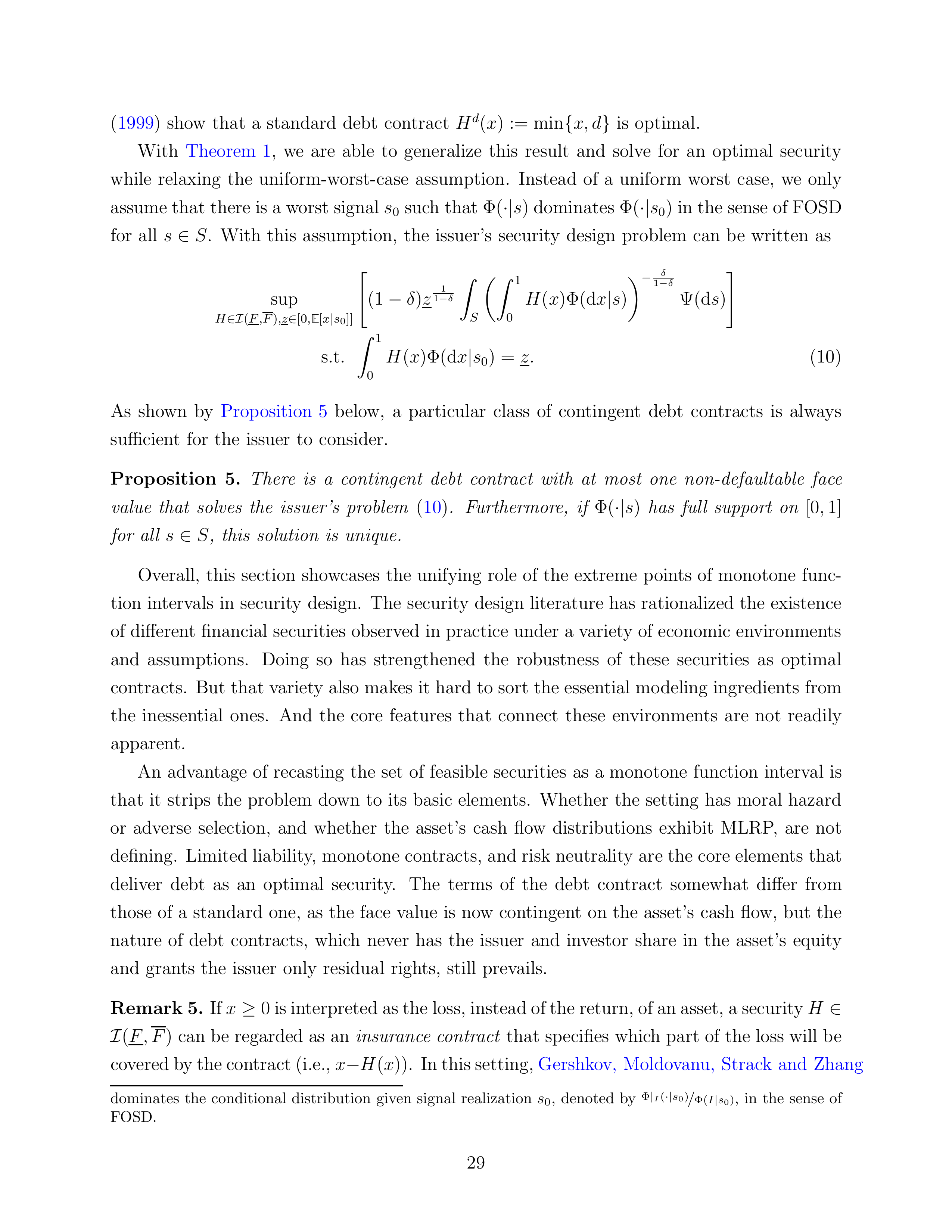}
                  & A contingent debt contract with at most one non-defaultable face value.\\
            \cmidrule{2-5}
                  & Repo-QA
                  & What does the Mastodon-hashtag-collector API method /hashtag/:hashtag return?
                  & \includegraphics[width=15mm,height=15mm]{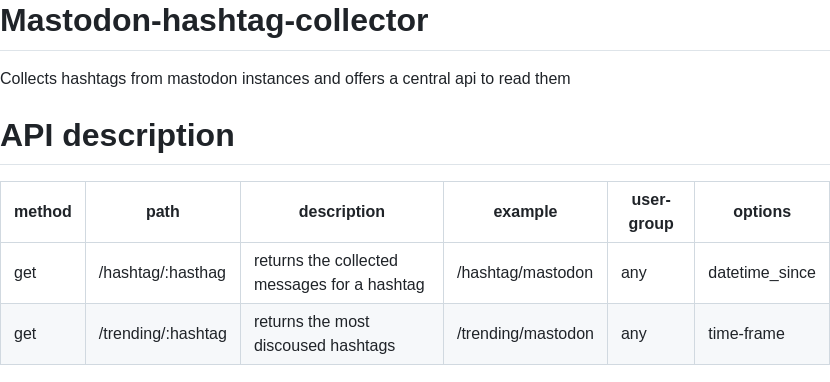}
                  & It returns the collected messages for a specific hashtag.\\
            \midrule
            \multirow{13}{*}{\makecell[c]{Open-Vocab\\ Classification}}
                  & \makecell[l]{Product\\ Classification}
                  & --
                  & \includegraphics[width=15mm,height=15mm]{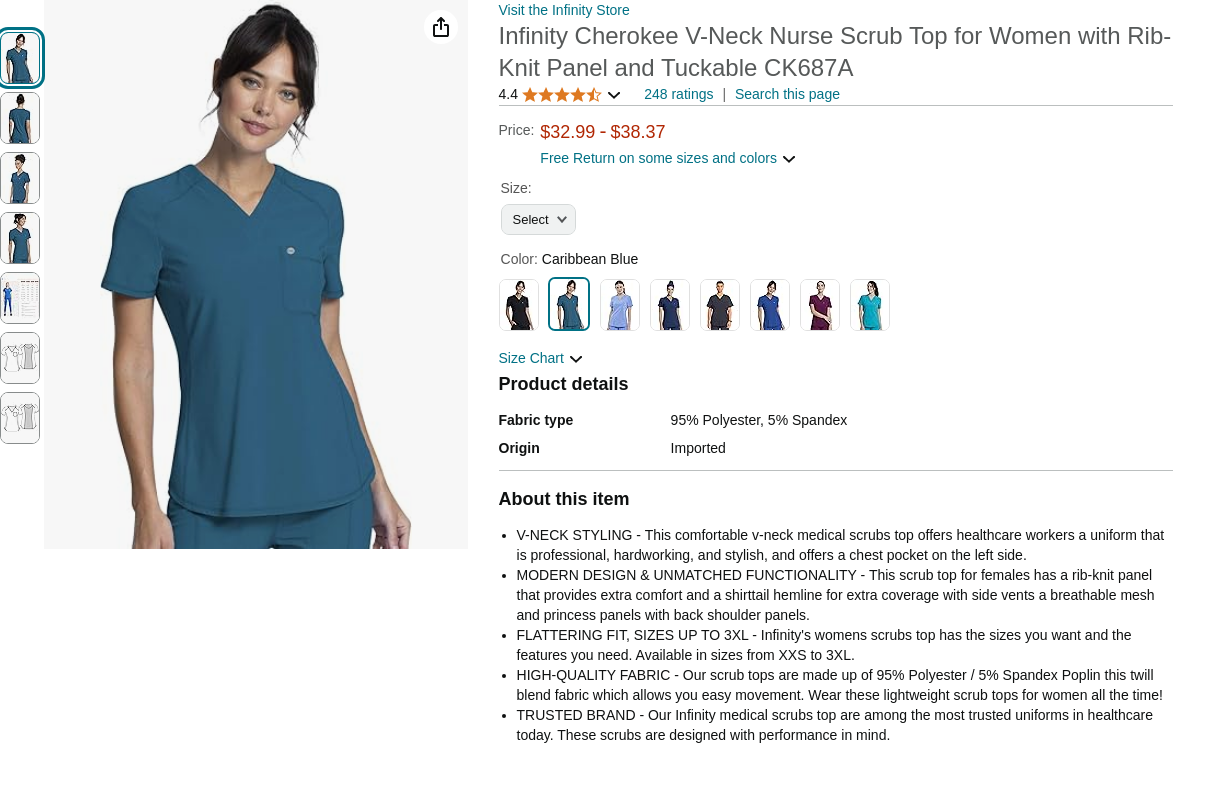}
                  & Clothing, Shoes \& Jewelry - Women - Uniforms, Work \& Safety - Clothing - Medical - Scrub Tops\\
            \cmidrule{2-5}
                  & News Topics
                  & --
                  & \includegraphics[width=15mm,height=15mm]{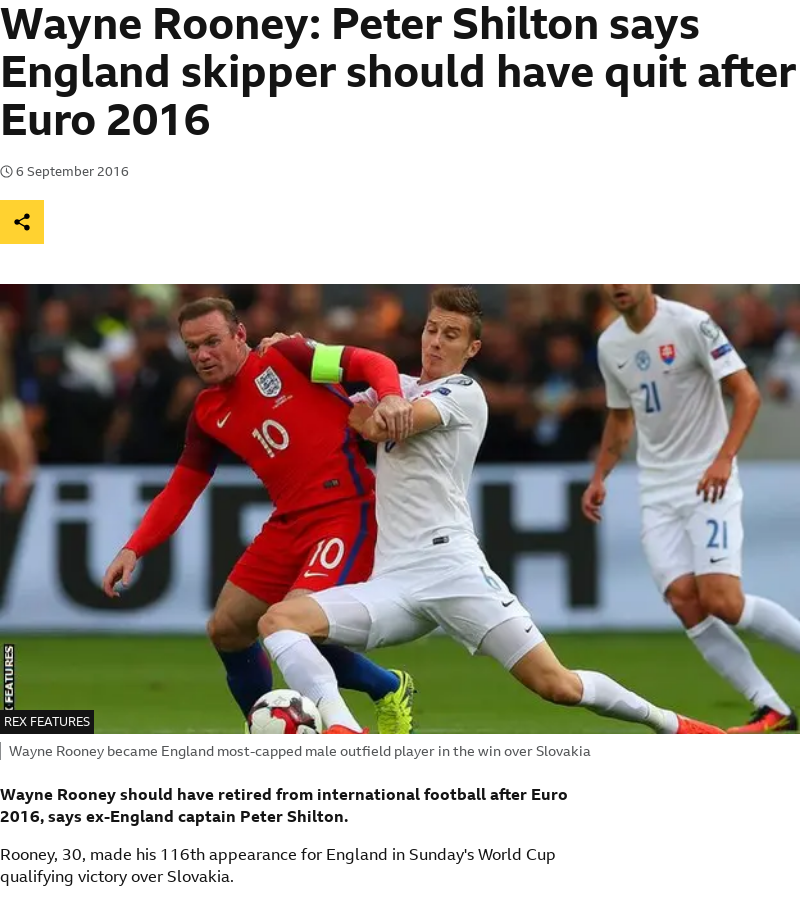}
                  & Sports, Football\\
            \cmidrule{2-5}
                  & Arxiv Fileds
                  & --
                  & \includegraphics[width=15mm,height=15mm]{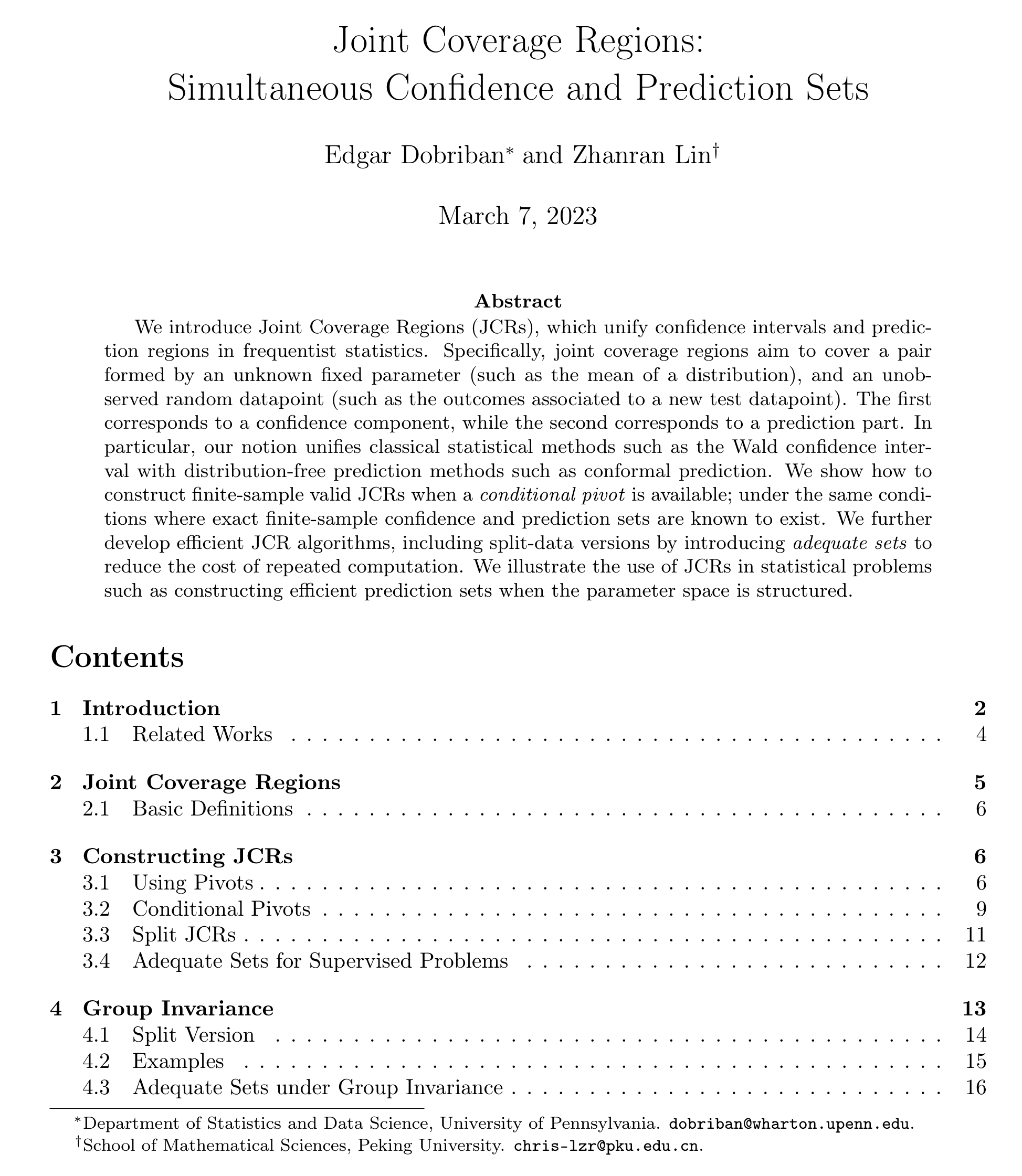}
                  & Artificial Intelligence\\
            \cmidrule{2-5}
                  & \makecell[l]{Knowledge\\ Classification}
                  & --
                  & \includegraphics[width=15mm,height=15mm]{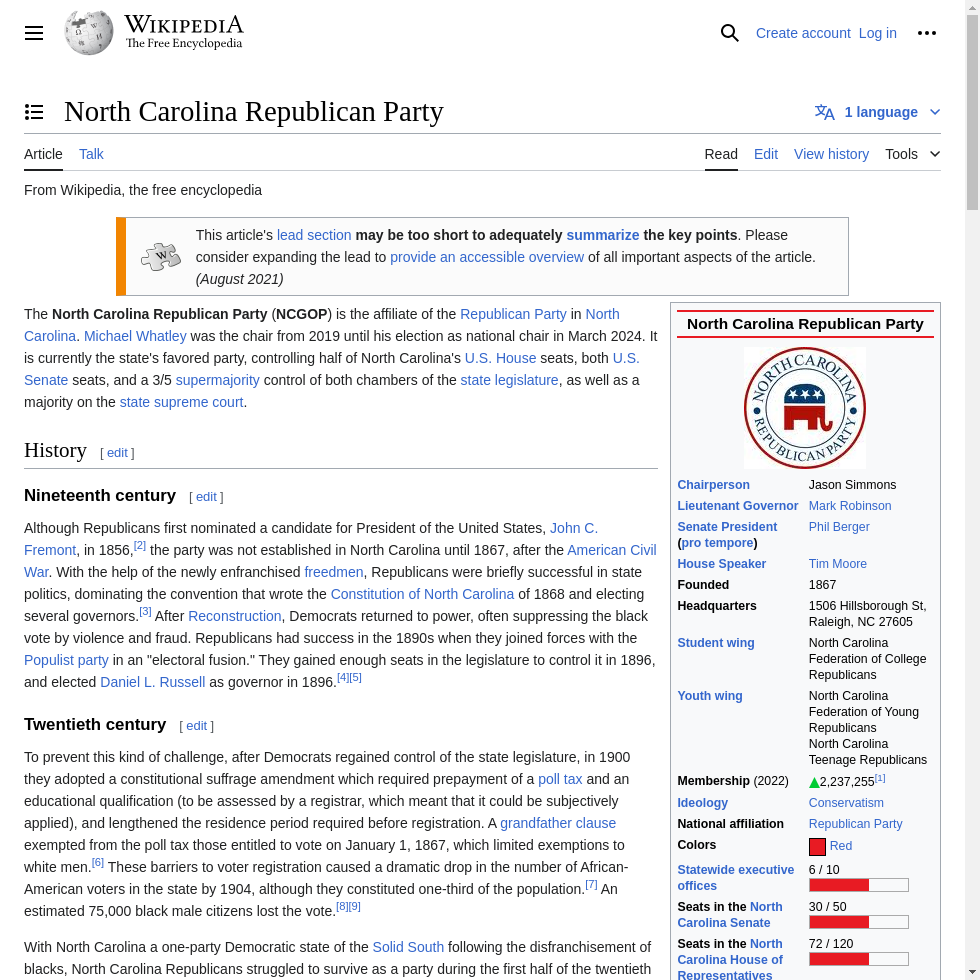}
                  & Society and social sciences, Government and Politics\\
            \bottomrule
    \end{tabular}
    \caption{Examples of tasks in Screenshot Question Answering and Open-Vocab Classification in MVRB.}
    \label{figure:MVRB-example-SQA-OVC}
    \end{center}
\end{figure*}

We construct our benchmark through machine annotation, human annotation, and filtering and reorganizing other publicly available datasets. Examples of tasks in our benchmark are presented in Figure~\ref{figure:MVRB-example-SR-CSR} and Figure~\ref{figure:MVRB-example-SQA-OVC}. The specific number of queries and the corpus for each subtask of our benchmark are listed in Table~\ref{tab:benchmark-data-statistics}. Below, we provide detailed information on our construction process for each meta task.

\subsection{Screenshot Retrieval}  
\begin{figure*}[tb]
    \centering
    \includegraphics[width=0.95\textwidth]{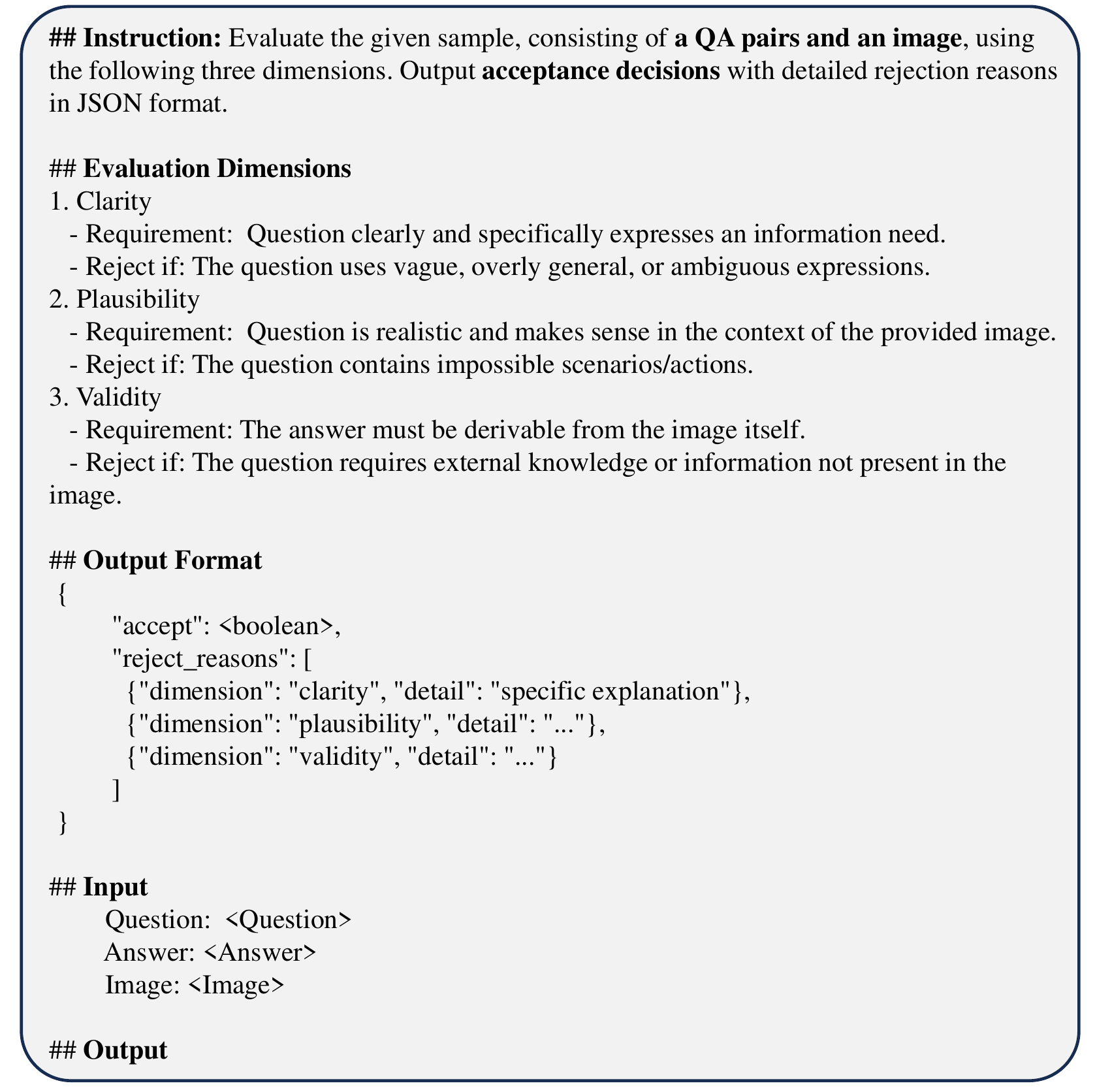}
    \vspace{-0.3cm}
    \caption{The prompt used for qulatiy control. We employ MLLMs to evaluate the clarity, plausibility, and validity of the evaluation examples.} 
    \vspace{-0.3cm}
    \label{fig:prompt-vlm-experts}
\end{figure*}

Screenshot Retrieval (SR) tasks are constructed through machine annotation and filtering/restructuring of publicly available datasets. Tasks constructed by machine annotation include \textit{product retrieval}, \textit{paper retrieval}, \textit{repo retrieval}, and \textit{news retrieval}. For these tasks, we use captions of screenshots to generate question-answer pairs using LLMs with the questions serving as queries. 
To ensure quality, we implement a two-stage quality control process involving \textbf{automatic assessment} and \textbf{human verification}. First, each sample is assessed by three MLLMs (LLaVA1.6-34B\footnote{https://huggingface.co/liuhaotian/llava-v1.6-34b}, Molmo-72B\footnote{https://huggingface.co/allenai/Molmo-72B-0924}, Llama-3.2-90B-Vision-Instruct\footnote{https://huggingface.co/meta-llama/Llama-3.2-90B-Vision-Instruct}) across three dimensions: 1) \textit{Clarity} : whether the query expresses a concrete information need, e.g., filtering out vague queries like "\textit{a sports car}", 2) \textit{Reasonableness}: whether the query is realistic, e.g., filtering out "\textit{a horse that can swim}"), and 3) \textit{Correctness}: whether the query can be answered by the retrieved screenshot, e.g., filtering out "\textit{what is the citation count of this paper}" since it cannot be answered solely from the screenshot, with the corresponding prompts detailed in Figure~\ref{fig:prompt-vlm-experts}. Each evaluation sample is independently reviewed by the MLLMs, and any sample failing any criterion is discarded. The remaining samples are further verified by human labelers using the same principles. An evaluation sample is successfully created only if it passes both stages of quality control.

Tasks derived from filtering and restructuring publicly available datasets include \textit{chart-retrieval}, \textit{document-retrieval}, and \textit{slide-retrieval}. We selected test sets from ChartVQA~\cite{chartvqa}, DocVQA~\cite{docvqa}, and SlideVQA~\cite{slidevqa}, using their questions as queries and corresponding images as target screenshots. To refine the evaluation dataset, we remove cases where a single query matches multiple images, such as in SlideVQA. We also filter out context-dependent questions, like "\textit{What is the content of Section 3.4.5?}", ensuring that each query is tailored for retrieving a single screenshot.

\begin{figure*}[tb]
    \centering
    \includegraphics[width=0.95\textwidth]{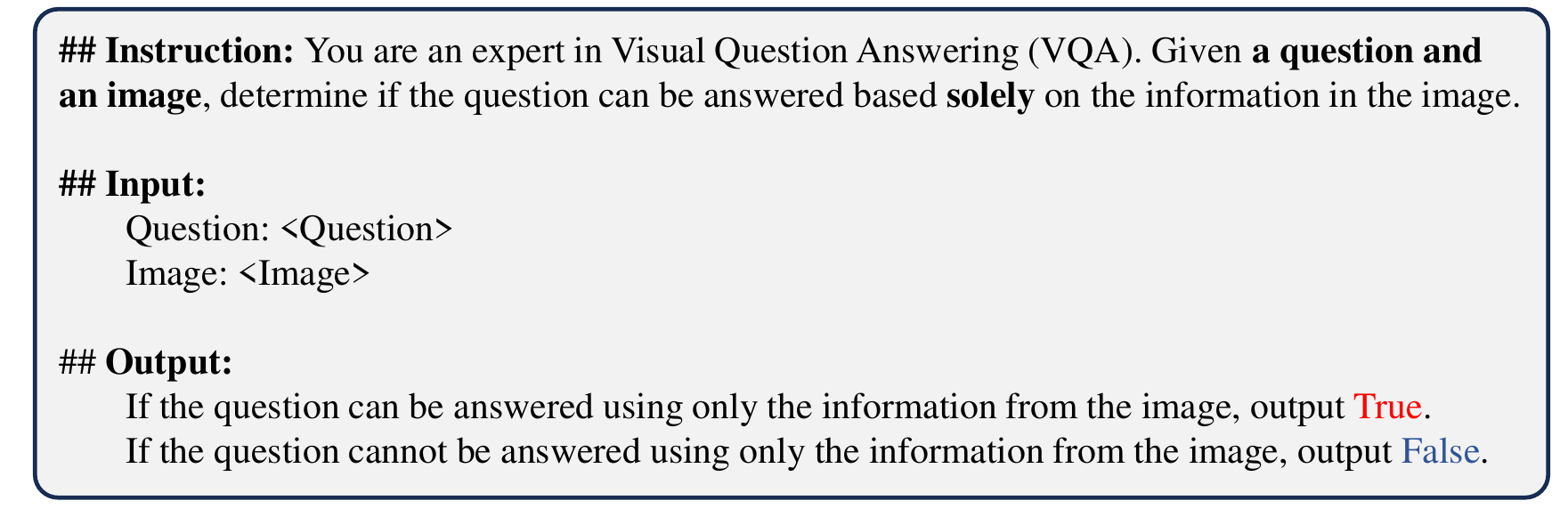}
    \vspace{-0.3cm}
    \caption{The prompt used for filtering out false negatives.} 
    \vspace{-0.3cm}
    \label{fig:prompt-vlm-false-negatives}
\end{figure*}

To increase task difficulty, we introduce hard negatives based on the query and target screenshot, following a method similar to the hard negatives augmentation approach for q2s tuples and sq2s triplets (Appendix~\ref{appendix:data-annotation}). To prevent false negatives, we use MLLM filtering to ensure no relevant screenshots are mistakenly included in the hard negatives, with the corresponding prompt shown in Figure~\ref{fig:prompt-vlm-false-negatives}.

\subsection{Composed Screenshot Retrieval}  
\label{appendix:CSR-details}
We construct these tasks through human annotation, encompassing four subtasks: \textit{product discovery}, \textit{news-to-Wiki}, \textit{knowledge relation}, and \textit{Wiki-to-product}. These tasks are not only challenging but also hold significant real-world relevance. Annotators are guided to gather query screenshots from appropriate platforms, generate realistic and high-quality queries, and acquire the matching target screenshots. To enhance both the complexity and the discriminative aspect of the task, each annotator also identifies 8$\sim$10 similar screenshots as difficult negatives for each original screenshot. Below, we outline the construction process for each specific CSR task:
\begin{itemize}
    \item \textbf{Product Discovery}: This task simulates a real-world shopping scenario where users search for a related product based on specific needs such as price, color, brand, or accessories. Annotators browse Amazon, search for products from a predefined category set, capture screenshots, and then generate queries based on the information in the screenshots. They then retrieve another product screenshot that satisfies the query.  

    \item \textbf{News-to-Wiki}: This task mirrors a scenario where users read news articles and seek additional details on people, events, locations, or causes, available in Wikipedia entries. Annotators explore news platforms like BBC, Fox, and CNN, develop queries from the content, and find relevant Wikipedia entries, capturing screenshots that answer the query.

    \item \textbf{Knowledge Relation}: This task simulates a scenario where users browsing Wikipedia are interested in exploring related topics. When looking at one entry, they often seek additional information, which is available in another Wikipedia article. Annotators search Wikipedia, capture relevant screenshots, formulate queries based on the screenshot content, and find another entry screenshot that fulfills the query. 

    \item  \textbf{Wiki-to-Product}: This task captures the scenario where users reading Wikipedia articles about products may wish to purchase them.  Annotators are given Wikipedia entries related to products and retrieve corresponding screenshots from Amazon and Wikipedia to create query-target pairs.
\end{itemize}

\subsection{Screenshot Question Answering}
\begin{figure*}[tb]
    \centering
    \includegraphics[width=0.95\textwidth]{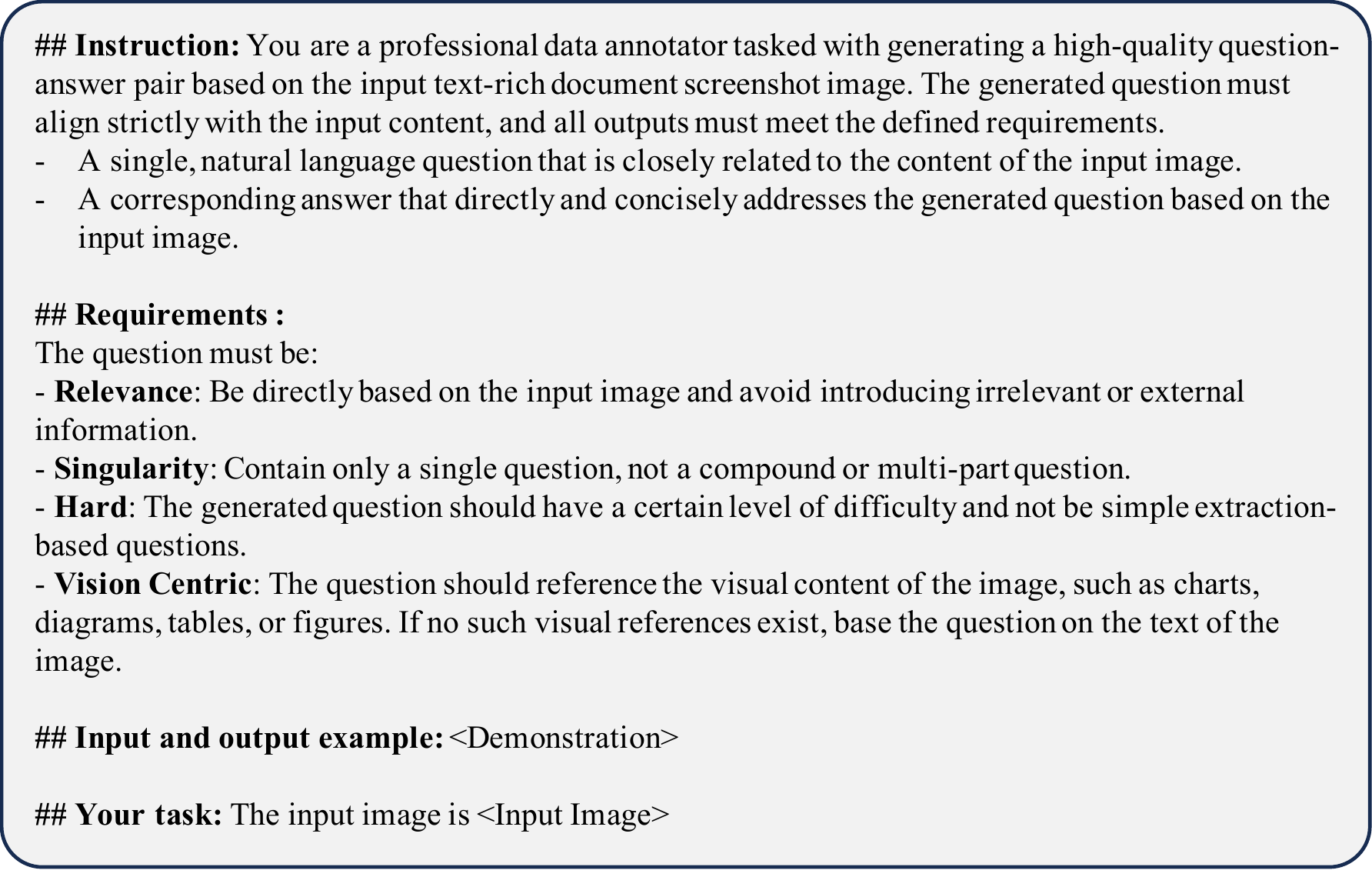}
    \vspace{-0.3cm}
    \caption{The prompt used for creating screenshot question answering task.} 
    \vspace{-0.3cm}
    \label{fig:prompt-vlm-sqa}
\end{figure*}

Screenshot Question Answering (SQA) tasks are constructed through machine annotation and include \textit{product-QA}, \textit{news-QA}, \textit{Wiki-QA}, \textit{paper-QA}, and \textit{repo-QA}. For these tasks, we utilize collected screenshots and generate question-answer pairs using MLLMs, with the corresponding prompts provided in Figure~\ref{fig:prompt-vlm-sqa}. To ensure quality of each evaluation example, we follow the same \textbf{automatic assessment} and \textbf{human verification} process used in creating SR tasks, assessing the evaluation example based on clarity, reasonableness, and correctness. To further increase task difficulty, we leverage large language models (LLMs) to generate hard negative answers—plausible yet incorrect answers that closely resemble the correct one—thereby making the task more challenging.

\subsection{Open-Vocab Classification} 
For Open-Vocab Classification (OVC) tasks, we start by constructing a mutually orthogonal and well-organized category corpus. Labels are then manually annotated on screenshots. The details of the taxonomy creation for each task are listed as follows:
\begin{itemize}
    \item \textbf{Product Classification}: Categories are sourced from Amazon product pages and use hierarchies like \textit{"Arts, Crafts \& Sewing - Knitting \& Crochet - Crochet Hooks"}. A word tree is used to refine these categories to ensure exclusivity.

    \item \textbf{News Topics}: Categories are collected from CNN, BBC, and Fox News and are manually consolidated to form a mutually exclusive and comprehensive set that covers a wide range of real-world scenarios.

    \item \textbf{Academic Fields}:  Predefined categories are obtained from the Arxiv homepage\footnote{https://arxiv.org/}.

    \item \textbf{Knowledge Classification}: Categories are defined based on the Wikipedia category page\footnote{https://en.wikipedia.org/wiki/Wikipedia:Contents/Categories} and are manually combined to create an exclusive and comprehensive classification set for Wikipedia entries.
\end{itemize}

\begin{table}[h]
\centering
\resizebox{\columnwidth}{!}{
\begin{tabular}{cccc}
\toprule
Task & SubTask & \#Query & \#Corpus \\
\midrule
\multirow{7}{*}{SR} 
& Product Retrieval& 496 & 5436 \\ \cmidrule{2-4}
& Paper Retrieval& 503 & 5021 \\ \cmidrule{2-4}
& Repo Retrieval&510 & 5024 \\ \cmidrule{2-4}
& News Retrieval& 491 & 5401 \\ \cmidrule{2-4}
& Chart Retrieval& 200 & 5000 \\ \cmidrule{2-4}
& Doc Retrieval& 200 & 5000 \\ \cmidrule{2-4}
& Slide Retrieval& 200 & 5000 \\ 
\cmidrule{1-4}
\multirow{4}{*}{CSR} 
& Product Discovery& 107 & 1012 \\ \cmidrule{2-4}
& News-to-Wiki & 101 & 1010 \\ \cmidrule{2-4}
& Knowledge Relation & 100 & 1011 \\ \cmidrule{2-4}
& Wiki-to-product & 100 & 969 \\ 
\cmidrule{1-4}
\multirow{5}{*}{SQA} 
& Product-QA & 498 & 2988 \\ \cmidrule{2-4}
& News-QA & 500 & 3000 \\ \cmidrule{2-4}
& Wiki-QA & 500 & 3000 \\ \cmidrule{2-4}
& Academic-QA & 481 & 2886 \\ \cmidrule{2-4}
& Repo-QA & 489 & 2934 \\
\cmidrule{1-4}
\multirow{4}{*}{OVC} 
& Product Classification & 500 & 5000 \\ \cmidrule{2-4}
& News Toptics & 584 & 74 \\ \cmidrule{2-4}
& Academic Fields & 500 & 142 \\ \cmidrule{2-4}
& Knowledge Classification & 480 & 46 \\
\bottomrule
\end{tabular}}
\caption{The specific number of queries and the corpus for each subtask}
\label{tab:benchmark-data-statistics}
\end{table}

\section{Baselines and Evaluation Setup}
\label{appendix-baselines}
We select retrieval models that are widely adopted in practice and academic research, which can be divided into three main categories: OCR + Text Retrievers, General Multimodal Retrievers, and Visualized Document Retrievers. In the OCR + Text Retrievers category, we used Paddle OCR to extract text from the image part and performed text concatenation in the order from the top-left to the bottom-right. For evaluation metrics, we used the Recall metric and recorded the Recall@1 for each task.

\subsection{OCR + Text Retrievers}
\label{appendix:baseline-ocr+text}
\noindent\textbf{BM25}~\cite{bm25-robertson2009probabilistic} is a classical information retrieval algorithm improved from TF-IDF, which optimizes relevance scoring through document length normalization (parameter $b$) and term frequency saturation mechanisms (parameter $k_1$). 
For evaluation, we use the code of \href{https://github.com/dorianbrown/rank_bm25}{dorianbrown/rank\_bm25}.

\noindent\textbf{DPR}~\cite{dpr} employs dual BERT encoders for dense retrieval, mapping queries and documents into a shared vector space via contrastive learning. Trained with an ``in-batch negatives'' strategy for efficiency. For evaluation, we use the weight of \href{https://huggingface.co/facebook/dpr-question_encoder-single-nq-base}{facebook/dpr-question\_encoder-single-nq-base}.

\noindent\textbf{BGE}~\cite{bge} is a widely used BERT-based text embedding model. It is trained on a large text corpus and optimized using contrastive learning. For evaluation, we utilize the base version of BGE with weights from \href{https://huggingface.co/BAAI/bge-base-en-v1.5}{BAAI/bge-base-en-v1.5}.

\noindent\textbf{E5-Mistral}~\cite{e5-mistral} is a large language model-based embedding model derived from the Mistral-7B architecture. It uses the last hidden state of the [EOS] token as the embedding for the input text. For evaluation, we utilize the weights from \href{https://huggingface.co/intfloat/e5-mistral-7b-instruct}{intfloat/e5-mistral-7b-instruct}.

\subsection{General Multimodal Retrievers}
\label{appendix:baseline-general}
\noindent\textbf{VISTA}~\cite{vista} is a universal multimodal retriever based on general text embedding models, such as pre-trained BERT models~\cite{bert-devlin2018bert,bge}. It processes input images by converting them into sequences of patches, which are then fed into the pre-trained text embedding model alongside text tokens in an interleaved manner. Although VISTA was trained on natural images, we evaluated its performance on the MVRB by directly inputting document screenshot images.

\noindent\textbf{Uni-IR}~\cite{mbeir-wei2023uniir} is a single retriever to accomplish any multimodal retrieval task. UniIR can follow the instructions to take a heterogeneous query to retrival from a heterogeneous candidate pool with millions of candidates in diverse modalities. We use the weight of \href{https://huggingface.co/TIGER-Lab/UniIR}{TIGER-Lab/UniIR}.

\noindent\textbf{CLIP}~\cite{clip-radford2021learning} is a multimodal contrastive learning model comprising a ViT image encoder and Transformer text encoder. Pretrained on 400 million image-text pairs, CLIP maps cross-modal content into a shared vector space, enabling zero-shot image classification and cross-modal retrieval. It processes an image as 14*14 patches and outputs the 1+196 (ClS token and 196 patch tokens) sequence to represent the image. We use \href{https://huggingface.co/openai/clip-vit-large-patch14}{openai/clip-vit-large-patch14} for evaluation.

\noindent\textbf{SigLIP}~\cite{siglip2023} is a multimodal model based on SoViT-400m~\cite{sovit400m} architecture with a better contrastive loss fuction. The sigmoid loss operates solely on image-text pairs and does not require a global view of the pairwise similarities for normalization. This allows further scaling up the batch size, while also performing better at smaller batch sizes. It pre-trained on WebLi~\cite{webli} at resolution 384*384. We use the weight of \href{https://huggingface.co/google/siglip-so400m-patch14-384}{google/siglip-so400m-patch14-384}.

\noindent\textbf{E5-V}~\cite{e5-v2024} is a unified multimodal embedding framework that maps text/image inputs into a shared semantic space via prompt mechanisms. Innovatively trained with text-only pairwise data through single-modality contrastive learning.  We use the weight of \href{https://huggingface.co/royokong/e5-v}{royokong/e5-v} for evaluation.

\noindent\textbf{VLM2Vec}~\cite{jiang2024vlm2vec} builds upon the multimodal large language model Phi-3.5-V~\cite{phi3}. For any given input, it leverages the last hidden state of the [EOS] token to generate embeddings. VLM2Vec was trained on 20 diverse multimodal datasets, encompassing a range of tasks such as classification, visual question answering, retrieval, and visual grounding. Although the primary focus of its training corpus is on natural images, it also includes several document question answering datasets, such as DocVQA~\cite{mathew2021docvqa}. This extensive training enables VLM2Vec to exhibit strong performance in the Screenshot Question Answering (SQA) tasks within our MVRB benchmark, outperforming other models, including our own UniSE, which are zero-shot for these tasks. Despite its competitive performance in SQA tasks due to this specialized training, VLM2Vec is classified as a general-purpose retriever in our taxonomy because of its limited exposure to document data, both in terms of quantity and the narrow focus on question answering tasks. For evaluation, we use the weight of \href{https://huggingface.co/TIGER-Lab/VLM2Vec-Full}{TIGER-Lab/VLM2Vec-Full}.

\noindent\textbf{MM-Embed}~\cite{nv-mm-embed2024} is a multimodal dual-encoder model addressing vision-text bias through modality-aware negative sample mining. It supports hybrid-modality queries (e.g., text+image combinations) and enhances retrieval precision in complex scenarios via large language model reranking. We use the weight of \href{https://huggingface.co/nvidia/MM-Embed}{nvidia/MM-Embed}.

\subsection{Screenshot Document Retrievers}
\noindent\textbf{DSE}~\cite{ma2024unifying} is a bi-encoder model designed to encode document screenshots into dense vectors for document retrieval. It trained on Wiki pages and documents. We use the weight of \href{https://huggingface.co/Tevatron/dse-phi3-docmatix-v2}{Tevatron/dse-phi3-docmatix-v2}. 

\noindent\textbf{ColPali}~\cite{faysse2024colpali} is an advanced document retrieval model designed to leverage vision-language models (VLMs) to generate high-quality contextual embeddings from document page images. The model is based on PaliGemma-3B~\cite{beyer2024paligemma} and employs the ColBERT~\cite{khattab2020colbert} strategy to create multi-vector representations of document page, thereby enhancing the accuracy and efficiency of retrieval. We use the weight of \href{https://huggingface.co/vidore/colpali-v1.3-hf}{vidore/colpali-v1.3-hf} for evaluation.

\noindent\textbf{GME}~\cite{zhang2024gme} is based on the Qwen2-VL model, and utilizes a contrastive learning approach to integrate diverse data into a unified semantic space. The GME has been trained on the visual document dataset Docmatix~\cite{laurenccon2024building}, which makes it have a great power to handle the visual document. For evaluation, we use the weight of \href{https://huggingface.co/Alibaba-NLP/gme-Qwen2-VL-2B-Instruct}{Alibaba-NLP/gme-Qwen2-VL-2B-Instruct}.

\vspace{0.5cm}

\newcounter{maintable} 
\renewcommand{\thetable}{\arabic{maintable} (\Alph{table})}

\setcounter{maintable}{7} 
\setcounter{table}{0} 
\begin{table*}[htp]
\centering
\resizebox{1.0\textwidth}{!}{
\begin{tabular}{lcccccccc}
\toprule
& \textbf{BM25} & \textbf{DPR} & \textbf{BGE} & \textbf{E5-Mistral} & \textbf{CLIP} & \textbf{SIGLIP} & \textbf{VISTA} & \textbf{Uni-IR} \\
\midrule
\rowcolor{gray!30} \textbf{SR} & & & & & & & &\\
Product          & 59.27 & 29.44 & 62.70 & 69.75 & 34.88 & 55.24 & 10.48  & 24.19 \\
Paper               & 70.38 & 23.66 & 50.89 & 69.18 & 6.96 & 26.24 & 0.60 & 1.19 \\
Repo       & 44.71 & 30.78 & 53.92 & 60.20 & 15.69 & 47.65 & 3.14 & 7.84 \\
News              & 38.08 & 26.48 & 43.18 & 47.45 & 24.24 & 33.20 & 4.28 & 11.20 \\
Chart              & 9.00 & 8.50 & 21.50 & 33.00 & 4.50 & 24.00 & 0.50 & 4.50 \\
Document   & 19.00 & 19.00 & 30.50 & 29.00 & 7.50 & 21.00 & 2.50 & 6.00 \\
Slide  & 44.50 & 37.00 & 57.00 & 51.50 & 38.50  & 61.00  & 15.00 & 31.50 \\
\textit{Avg} & 40.71 & 24.98 & 45.67 & 51.44 & 18.89 & 38.33 & 5.21 & 12.35 \\
\midrule

\rowcolor{gray!30} \textbf{CSR} & & & & & & & & \\
Knowledge Relation               & 19.00  & 15.00 & 18.00 & 23.00 & 18.00 & 30.00 & 6.00 & 23.00 \\
News2Wiki              & 20.79 & 10.89 & 17.82 & 31.68 & 16.83 & 21.78 & 6.93 & 18.81 \\
Product Discovery               & 25.23 & 28.04 & 27.10 & 33.64 & 32.71  & 55.14 & 25.23 & 29.91 \\
Wiki2Product      & 49.00 & 21.00 & 84.00 & 92.00 & 34.00 & 31.00 & 7.00 & 47.00 \\
\textit{Avg}      & 28.51  & 18.73 & 36.73 & 45.08 & 25.39  & 34.48 & 11.29 & 29.68 \\
\midrule

\rowcolor{gray!30} \textbf{SQA} & & & & & & & & \\
Repo-QA              & 46.63 & 35.38 & 49.69 & 33.33 & 27.81 & 24.54 & 37.63 & 29.24 \\
News-QA                 & 38.40 & 35.40 & 38.80 & 57.60  & 31.80 & 19.80 & 24.00 & 22.20 \\
Product-QA & 22.49 & 10.04 & 26.31 & 27.71 & 15.26 & 16.06 & 15.46 & 12.85 \\
Paper-QA      & 46.78 & 30.15 & 32.85 & 52.18 & 22.45 & 21.62 & 31.60 & 21.62 \\
Wiki-QA          & 38.00 & 25.80 & 33.80 & 44.20 & 22.20 & 16.00 & 20.20 & 21.20 \\
\textit{Avg} & 38.46 & 27.25 & 36.29 & 43.00 & 23.90 & 19.60 & 25.78 & 21.42 \\
\midrule

\rowcolor{gray!30} \textbf{OVC} & & & & & & & & \\
Product Classification              & 7.6  & 1.60 & 36.20 & 41.00 & 21.00 & 39.40 & 6.80 & 10.40 \\
News  Topics            & 8.22  & 39.04  & 63.35 & 59.42 & 50.68 & 63.18 & 30.99 & 39.38 \\
Knowledge Classification               & 7.08  & 37.08 & 52.50 & 42.08 & 42.91  & 44.58 & 26.25 & 26.67 \\
Academic Fileds      & 2.00 & 10.60 & 39.60 & 12.20 & 7.00 & 15.40 & 2.40 & 3.80 \\
\textit{Avg}      & 6.23  & 22.08 & 47.91 & 38.68 & 30.40  & 40.64 & 16.61 & 20.06 \\
\midrule

\rowcolor{gray!30} \textbf{Overall Score} & & & & &  & & & \\
All                  & 30.81 & 23.74 & 41.99 & 45.51  & 23.75 & 33.34 & 13.85 & 19.63 \\
\bottomrule
\end{tabular}
}
\caption{Detailed performance results for each task of different retrievers on MVRB (Recall@1).}
\label{tab:main_exp_per_task}
\end{table*}

\setcounter{table}{1}
\begin{table*}[htp]
\centering
\resizebox{1.0\textwidth}{!}{
\begin{tabular}{lcccccccc}
\toprule
& \textbf{VLM2Vec} & \textbf{E5-V} & \textbf{MM-Embed} & \textbf{DSE} & \textbf{Colpali} & \textbf{GME} & \textbf{UniSE-CLIP} & \textbf{UniSE-MLLM} \\
\midrule
\rowcolor{gray!30} \textbf{SR} & & & & & & & &\\
Product          & 23.59 & 33.46 & 28.83 & 71.17 & 72.50 & 57.80 & 53.02 & 75.40 \\
Paper              & 7.75 & 39.36 & 18.09 & 78.02 & 80.31 & 74.55 & 34.19 & 90.26 \\
Repo         & 17.65 & 36.08 & 30.59 & 74.01 & 68.23 & 76.47 & 40.98 & 83.33 \\
News              & 17.52 & 26.88 & 24.03 & 50.61 & 41.10 & 57.00 & 39.92 & 75.96 \\
Chart               & 11.50 & 23.00 & 11.00 & 29.00 & 24.00 & 39.00 & 11.00 & 36.50 \\
Document  & 12.00 & 26.00 & 18.50 & 47.50 & 59.50 & 51.00 & 17.00 & 50.50 \\
Slide & 21.50 & 54.00 & 50.00 & 80.50  & 86.50  & 75.50 & 55.50 & 75.50 \\
\textit{Avg} & 15.93 & 34.11 & 25.86 & 61.54 & 61.73 & 61.62 & 35.95 & 69.64 \\
\midrule

\rowcolor{gray!30} \textbf{CSR} & & & & & & & & \\
Knowledge Relation             & 47.00  & 18.00  & 26.00  & 22.92  & 16.00  & 20.00 & 29.00 &  41.00 \\
News2Wiki            & 49.50  & 37.62  & 41.58  & 21.88  & 16.80    & 19.80 & 27.72 & 44.55 \\
Product Discovery             & 32.71  & 34.58  & 41.12 & 24.04  & 25.20    & 28.90 & 45.79 & 51.40 \\
Wiki2Product      & 63.00 & 30.00 & 55.00 & 82.29  & 82.00   & 82.00 & 71.00 & 81.00\\
\textit{Avg}    & 48.05 & 30.05 & 40.93 & 37.78  & 35.00   & 37.68 & 43.38 & 54.49 \\
\midrule

\rowcolor{gray!30} \textbf{SQA} & & & & & & & & \\
Repo-QA            & 58.28 & 36.81 & 53.99  & 46.93 & 42.33  & 48.80 & 35.58 & 57.00 \\
News-QA               & 51.40 & 36.20 & 50.20  & 40.52 & 40.40  & 43.00 & 33.00 & 47.60 \\
Product-QA     & 36.55 & 19.68 & 30.12 & 22.98 & 17.67  & 18.80 & 19.68 & 25.10 \\
Paper-QA     & 43.78 & 34.10 & 38.46 & 46.67 & 46.98  & 45.30 & 25.36 & 46.70 \\
Wiki-QA        & 57.00 & 30.20 & 41.40  & 39.11 & 29.20  & 33.00 & 27.00 & 39.60 \\
\textit{Avg}& 49.42 & 31.40 & 42.83 & 39.24 & 35.32  & 37.78 & 28.13 & 43.20 \\
\midrule

\rowcolor{gray!30} \textbf{OVC} & & & & & & & & \\
Product Classification         & 20.80 & 14.20 & 26.80 & 30.00 & 29.60 & 33.40 & 33.20 & 45.40 \\
News Topics & 46.75 & 52.57 & 46.75 & 56.68 & 44.00 & 69.60 & 57.19 & 69.34 \\
Knowledge Classification   & 14.79 & 45.42 & 43.13 & 23.54 & 33.95 & 58.50 & 52.29 & 52.91 \\
Academic Fileds & 10.60 & 19.20 & 14.00 & 15.80 & 16.60 & 30.40 & 19.80 & 25.40 \\
\textit{Avg} & 23.24 & 32.85 & 32.67 & 31.51 & 31.04 & 47.98 & 40.62 & 48.26 \\
\midrule

\rowcolor{gray!30} \textbf{Overall Score} & & & & &  & & & \\
All                 & 32.19 & 32.37  & 34.48 & 48.14 & 43.64  & 13.3 & 34.99 & 55.72 \\
\bottomrule
\end{tabular}
}
\caption{Detailed performance results for each task of different retrievers on MVRB (Recall@1).}
\label{tab:main_exp_per_task2}
\end{table*}

\section{Detailed Results for Each Task}
\label{appendix-detailed results}
The detailed results of different retrievers for each task  on MVRB are presented in Table~\ref{tab:main_exp_per_task} and Table~\ref{tab:main_exp_per_task2}.

\end{document}